\documentclass{article}

\usepackage{arxiv}

\usepackage[utf8]{inputenc} 
\usepackage[T1]{fontenc}    
\usepackage{hyperref}       
\usepackage{url}            
\usepackage{booktabs}       
\usepackage{amsfonts}       
\usepackage{nicefrac}       
\usepackage{microtype}      
\usepackage{lipsum}		
\usepackage{subcaption} 
\usepackage{graphicx}
\usepackage[numbers]{natbib}  

\usepackage{algorithm}
\usepackage{algpseudocode}

\usepackage{amsmath}
\usepackage{amssymb}

\usepackage{placeins}  

\usepackage{tikz}
\usepackage{multirow}
\usetikzlibrary{shapes, arrows.meta, positioning, fit, backgrounds, calc}
\usepackage{doi}

\usepackage{subcaption}
\usepackage{float}

\title{Intrinsic-Metric Physics-Informed Neural Networks (IM-PINN) for Reaction-Diffusion Dynamics on Complex Riemannian Manifolds}

\author{
\begin{tabular}{cc}
    \begin{minipage}{0.45\textwidth}
        \centering
        \href{https://orcid.org/0009-0005-4657-6472}{\includegraphics[scale=0.06]{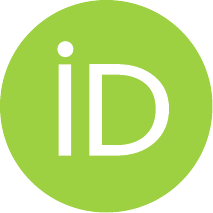}\hspace{1mm}Julian Evan Chrisnanto}\thanks{Corresponding author.} \\
        \normalfont
        Department of Bio-Functions and Systems Science\\
        Graduate School of Bio-Applications and Systems Engineering\\
        Tokyo University of Agriculture and Technology\\
        2-24-16 Nakacho, Koganei, Tokyo 184-8588, Japan \\
        \texttt{s254167v@st.go.tuat.ac.jp}
    \end{minipage}
    &
    \begin{minipage}{0.45\textwidth}
        \centering
        \href{https://orcid.org/0009-0008-0761-1249}{\includegraphics[scale=0.06]{orcid.pdf}\hspace{1mm}Salsabila Rahma Alia} \\
        \normalfont
        Department of Mathematics\\
    	Faculty of Mathematics and Natural Sciences\\
    	Universitas Padjadjaran \\
        Jl. Raya Bandung-Sumedang KM21, Jatinangor, Sumedang, West Java, Indonesia\\
    	\texttt{salsabila19026@mail.unpad.ac.id} 
    \end{minipage}
    \\[10ex]
    \begin{minipage}{0.45\textwidth}
        \centering
        Nurfauzi Fadillah \\
        \normalfont
        PLABS.ID\\
        Jl. Batununggal Mulia IV No.16, \\
        Bandung, West Java 40267, Indonesia \\
        \texttt{fauzi@plabs.id}
    \end{minipage}
    &
    \begin{minipage}{0.45\textwidth}
        \centering
        \href{https://orcid.org/0000-0002-3491-3049}{\includegraphics[scale=0.06]{orcid.pdf}\hspace{1mm}Yulison Herry Chrisnanto} \\
        \normalfont
        Department of Informatics\\
        Faculty of Science and Informatics \\
        Jenderal of Achmad Yani University \\
        Jl. Terusan Jenderal Sudirman, Cimahi, West Java 40531, Indonesia \\
        \texttt{yhc@if.unjani.ac.id}
    \end{minipage}
\end{tabular}
}

\renewcommand{\shorttitle}{Intrinsic-Metric Physics-Informed Neural Networks (IM-PINN) for Reaction-Diffusion ...}

\hypersetup{
pdftitle={A template for the arxiv style},
pdfsubject={q-bio.NC, q-bio.QM},
pdfauthor={David S.~Hippocampus, Elias D.~Striatum},
pdfkeywords={First keyword, Second keyword, More},
}

\begin{document}
\maketitle

\begin{abstract}
Simulating nonlinear reaction-diffusion dynamics on complex, non-Euclidean manifolds remains a fundamental challenge in computational morphogenesis, constrained by high-fidelity mesh generation costs and numerical drift in discrete time-stepping schemes. This study introduces the Intrinsic-Metric Physics-Informed Neural Network (IM-PINN), a mesh-free geometric deep learning framework that solves partial differential equations directly in the continuous parametric domain. By embedding the Riemannian metric tensor into the automatic differentiation graph, our architecture analytically computes the Laplace-Beltrami operator, decoupling solution complexity from geometric discretization. We validate the framework on a "Stochastic Cloth" manifold with extreme Gaussian curvature fluctuations ($K \in [-2489, 3580]$), where traditional adaptive refinement fails to resolve anisotropic Turing instabilities. Using a dual-stream architecture with Fourier feature embeddings to mitigate spectral bias, the IM-PINN recovers the "splitting spot" and "labyrinthine" regimes of the Gray-Scott model. Benchmarking against the Surface Finite Element Method (SFEM) reveals superior physical rigor: the IM-PINN achieves global mass conservation error of $\mathcal{E}_{mass} \approx 0.157$ versus SFEM's $0.258$, acting as a thermodynamically consistent global solver that eliminates mass drift inherent in semi-implicit integration. The framework offers a memory-efficient, resolution-independent paradigm for simulating biological pattern formation on evolving surfaces, bridging differential geometry and physics-informed machine learning.
\end{abstract}


\section{Introduction}

Spatiotemporal pattern formation via reaction-diffusion mechanisms constitutes a cornerstone of mathematical biology and non-equilibrium thermodynamics, providing a rigorous explanation for self-organizing phenomena ranging from animal coat markings to chemical oscillators \cite{Turing1952TheCBO, Gray1984AutocatalyticRI}. While the theoretical underpinnings of these systems are well-established on planar Euclidean domains, physical reality often necessitates their study on complex, non-Euclidean geometries—such as biological membranes, developing embryos, or functionalized textile surfaces. On such Riemannian manifolds, the diffusion process is no longer isotropic in the embedding space but is strictly governed by the intrinsic surface geometry, necessitating the replacement of the standard Laplacian with the Laplace-Beltrami operator \cite{Dziuk2013FiniteEM}. This geometric coupling introduces non-trivial dependencies where local curvature modulates transport efficiency and symmetry breaking, fundamentally altering the resulting equilibrium states compared to their flat counterparts \cite{Bronstein2017GeometricDL}.

The numerical approximation of these intrinsic operators has traditionally relied on grid-based discretization schemes, most notably the Surface Finite Element Method (SFEM) \cite{Dziuk2013FiniteEM}. While mathematically robust for static and smooth geometries, SFEM imposes a severe computational bottleneck when applied to the complex, stochastically perturbed manifolds characteristic of realistic physical modeling. Specifically, faithfully resolving the multi-scale features—such as the high-frequency wrinkles and random Gaussian perturbations inherent to a textile surface—requires the generation of an extremely dense and high-quality triangular tessellation. This meshing process is not only computationally expensive but algorithmically brittle; the emergence of degenerate elements (slivers) on high-curvature folds often leads to ill-conditioned stiffness matrices and numerical divergence, necessitating sophisticated and fragile remeshing subroutines. Crucially, this limitation extends to the majority of contemporary geometric deep learning paradigms; state-of-the-art architectures, including Simplicial Neural Networks \cite{Choi2024SNNPDELDF} and Mixture Model CNNs \cite{Monti2017GeometricDL}, ultimately remain bound to a discrete representation of the domain, relying on fixed connectivity graphs or simplicial complexes to define convolution. Consequently, in scenarios involving parametric uncertainty where the manifold geometry itself is a random variable requiring Monte Carlo sampling, the necessity of repeated, error-prone discretization renders these discrete frameworks computationally intractable, thereby motivating the search for a fundamentally continuous and mesh-free alternatives.

In response to these discretization challenges, the paradigm of Physics-Informed Neural Networks (PINNs) has emerged as a transformative mesh-agnostic framework for solving partial differential equations (PDEs), leveraging the universal approximation capabilities of deep neural networks to parameterize the solution directly in continuous space \cite{Raissi2019PhysicsInformedNN}. By embedding the governing physical laws—specifically the PDE residuals and boundary conditions—directly into the loss function via automatic differentiation, PINNs theoretically circumvent the need for tessellation and are naturally suited for inverse problems and irregular domains. However, the naive application of standard Euclidean PINN architectures to the study of reaction-diffusion dynamics on Riemannian manifolds is fraught with theoretical and practical perils. As elucidated by recent critical studies on the failure modes of physics-informed learning, standard PINNs exhibit severe spectral bias and convergence pathologies when applied to stiff, multi-scale systems like the Gray-Scott model, where the optimization landscape is marred by sharp local minima and vanishing gradients \cite{Krishnapriyan2021CharacterizingPF}. Furthermore, the extension of this framework to curved surfaces is non-trivial; it requires the replacement of the standard Laplacian with the geometric Laplace-Beltrami operator, a modification that introduces high-order derivative terms involving the metric tensor and its determinant, thereby exacerbating the bias-variance trade-off inherent in high-dimensional gradient estimation \cite{Hu2023BiasVarianceTI}. Consequently, while standard optimization strategies may suffice for simple elliptic problems, they frequently fail to capture the delicate symmetry-breaking bifurcations required for Turing pattern formation, necessitating a specialized architectural approach that rigorously respects both the symplectic structure of the dynamics and the intrinsic curvature of the domain \cite{Bihlo2023ImprovingPN}.

In this work, we present the Intrinsic-Metric Physics-Informed Neural Network (IM-PINN), a unified framework designed to resolve these discretization bottlenecks by solving the Gray-Scott reaction-diffusion equations directly on stochastically perturbed, cloth-like 3D manifolds. Unlike graph-based approaches that approximate geometry via adjacency matrices, IM-PINN treats the domain as a continuous Riemannian manifold encoded strictly via the metric tensor, thereby preserving the exact differential structure of the governing equations. This approach situates itself at the intersection of several rapidly advancing frontiers in geometric deep learning. While extensive recent surveys have cataloged the explosion of geometric optimization techniques \cite{Fei2023ASOB} and their applications to tasks ranging from molecular representation \cite{Atzt2021GeometricDL} and industrial cyber-physical systems \cite{VillalbaDiez2020GeometricDL} to time-series modeling \cite{Jeong2023DeepEC}, the majority of these efforts focus on static representation learning \cite{Ansuini2019IntrinsicDO, Meng2024StateRL} or latent manifold flattening \cite{Chen2020LearningFL}. Similarly, while significant strides have been made in adapting fundamental neural components—such as Riemannian Batch Normalization \cite{Brooks2019RiemannianBNA}, trivialization strategies \cite{Casado2019TrivializationsFGC}, and multinomial logistic regression \cite{Chen2024RMLRE}—to non-Euclidean settings, these methods typically address classification or regression rather than the distinct challenge of evolving dynamics governed by anisotropic differential operators. We further distinguish our contribution from the emerging family of operator learning methods, such as Phase-Field DeepONets \cite{Li2023PhaseFieldD} and Neural Operators \cite{Pert2024ScalingFS}, as well as projection-based Reduced Order Models \cite{Sibuet2025DiscretePT, Patsatzis2023SlowIM}; while these approaches offer computational speedups for parametrized families of equations, they rely on extensive offline data generation which is computationally prohibitive for the arbitrary, high-frequency curvature variations characteristic of our "Stochastic Cloth." By leveraging insights from manifold stochastics \cite{Sommer2021HorizontalFAM, Orlova2023DeepSMH} and addressing the "multiple manifold" complexity inherent in high-dimensional data \cite{Buchanan2021DeepNA}, the IM-PINN constructs a solver that is robust to geometric noise without requiring training data. Moreover, we incorporate hard physical constraints to mitigate optimization bias \cite{Lu2021PhysicsInformedNN} and draw a clear boundary with recent generative approaches like Geometry-Informed Neural Networks (GINNs) \cite{Berzins2025GeometryInformedNN}, which focus on generating static shapes satisfying geometric constraints rather than simulating time-dependent physics upon them. While we acknowledge the potential of frontier techniques like Geometric Clifford Algebra Networks \cite{Ruhe2023GeometricCAG} and adaptive log-Euclidean metrics \cite{Chen2023AdaptiveLMP}, our work establishes the fundamental baseline for intrinsic PDE solving, validated against competitive benchmarks and exploring novel curvature-fairness relationships \cite{Ma2024PredictingAE}.

The remainder of this manuscript is meticulously structured to establish both the theoretical rigor and the computational utility of the proposed framework. \textbf{Section 2 (Methodology)} provides the foundational mathematical derivation of the "Stochastic Cloth," explicitly defining the parametric height functions that induce multi-scale Gaussian perturbations. Here, we formulate the reaction-diffusion dynamics via intrinsic differential operators, detailing the construction of the Riemannian metric tensor $g_{ij}$ and the corresponding "Dual-Stream" neural architecture equipped with Fourier feature embeddings and the mass-conservative loss function. \textbf{Section 3 (Results)} presents a comprehensive empirical evaluation, benchmarking the IM-PINN against the SFEM \cite{Dziuk2013FiniteEM} across a battery of quantitative metrics, including pointwise tracking error, manifold capacity, and pattern topology verification. \textbf{Section 4 (Discussion)} synthesizes these findings, offering a phenomenological analysis of how intrinsic curvature modulates Turing instability and addressing the stability concerns raised in recent physics-informed learning literature \cite{Krishnapriyan2021CharacterizingPF} through the lens of thermodynamic consistency. Finally, \textbf{Section 5 (Conclusion)} summarizes the transformative potential of mesh-free geometric deep learning and outlines strategic avenues for future research, particularly regarding time-evolving domains and inverse shape design.

\section{Methods}
\subsection{Construction of the Stochastic Cloth Manifold}
We define the computational domain as a two-dimensional Riemannian manifold $\mathcal{M}$ embedded in $\mathbb{R}^3$, constructed to emulate the complex, multi-scale topography of a stochastically perturbed textile surface. Let $\Omega = [0,1]^2 \subset \mathbb{R}^2$ denote the parametric coordinate space with coordinates $\xi = (u,v)$. The manifold is realized via a smooth immersion $\mathbf{r}: \Omega \to \mathbb{R}^3$, given explicitly by $\mathbf{r}(u,v) = [u, v, z(u,v)]^\top$. To capture the phenomenological characteristics of a non-idealized physical membrane, the height function $z(u,v)$ is modeled not as a deterministic scalar field but as a realization of a random field comprising a macroscopic structural component and a microscopic stochastic perturbation. Specifically, we synthesize the surface elevation as a superposition of multi-frequency sinusoidal wrinkles and a Gaussian Random Field (GRF), formulated as:
\begin{equation}
    z(u,v) = \underbrace{\sum_{k=1}^{K} A_k \sin(2\pi \omega_{u,k} u + \phi_k) \cos(2\pi \omega_{v,k} v)}_{\text{Deterministic Wrinkles}} + \underbrace{\sigma \mathcal{G}(u,v; \ell)}_{\text{Stochastic Fluctuation}} + \mathcal{S}_{sag}(u,v),
\end{equation}
where $A_k$ denotes the amplitude of the $k$-th wrinkle mode, $\mathcal{G}$ represents a Gaussian process with correlation length $\ell$ governing the surface roughness \cite{Sommer2021HorizontalFAM}, and $\mathcal{S}_{sag}$ accounts for gravitational sagging. This construction induces a non-Euclidean metric structure on $\mathcal{M}$, completely characterized by the First Fundamental Form, or the metric tensor $\mathbf{g} = [g_{ij}] \in \mathbb{R}^{2 \times 2}$. The components of this symmetric positive definite (SPD) tensor are derived analytically as $g_{ij} = \langle \partial_i \mathbf{r}, \partial_j \mathbf{r} \rangle_{\mathbb{R}^3}$, where $\partial_1 = \partial/\partial u$ and $\partial_2 = \partial/\partial v$. Explicitly, the metric tensor takes the form:
\begin{equation}
    g_{ij}(u,v) = \delta_{ij} + (\partial_i z)(\partial_j z),
\end{equation}
where $\delta_{ij}$ is the Kronecker delta. The resultant geometry possesses a spatially varying determinant $|g| = \det(\mathbf{g}) = 1 + (\partial_u z)^2 + (\partial_v z)^2$, which acts as the local area element and is crucial for defining the intrinsic integration measure $d\mathcal{M} = \sqrt{|g|} du dv$ required for mass conservation \cite{Dziuk2013FiniteEM}. By retaining the full metric information, our formulation avoids the distortion artifacts inherent in conformal flattening approaches \cite{Chen2020LearningFL} and provides the necessary geometric priors for the subsequent reaction-diffusion analysis.

\subsection{Intrinsic Reaction-Diffusion Dynamics}
The spatiotemporal evolution of the chemical species on the manifold $\mathcal{M}$ is governed by the Gray-Scott reaction-diffusion system, a canonical model for pattern formation in non-equilibrium thermodynamics \cite{Gray1984AutocatalyticRI}. Unlike planar formulations, the transport of the species concentrations $U(\xi, t)$ and $V(\xi, t)$ is dictated by the intrinsic curvature of the domain. The coupled system of partial differential equations is given by:
\begin{align}
    \frac{\partial U}{\partial t} &= D_u \Delta_{\mathcal{M}} U - UV^2 + F(\xi)(1-U), \\
    \frac{\partial V}{\partial t} &= D_v \Delta_{\mathcal{M}} V + UV^2 - (F(\xi)+k)V,
\end{align}
where $D_u, D_v$ are the diffusion coefficients, $k$ is the dimensionless removal rate, and $F(\xi)$ is the dimensionless feed rate. Crucially, the diffusion term is mediated by the Laplace-Beltrami operator $\Delta_{\mathcal{M}}$, which generalizes the Euclidean Laplacian to Riemannian manifolds \cite{Turing1952TheCBO}. In our local parametric coordinates $\xi=(u,v)$, this operator is explicitly defined via the divergence of the gradient with respect to the metric tensor $\mathbf{g}$:
\begin{equation}
    \Delta_{\mathcal{M}} \psi = \text{div}_{\mathcal{M}}(\nabla_{\mathcal{M}} \psi) = \frac{1}{\sqrt{|g|}} \sum_{i,j=1}^2 \frac{\partial}{\partial \xi^i} \left( \sqrt{|g|} g^{ij} \frac{\partial \psi}{\partial \xi^j} \right),
\end{equation}
where $g^{ij} = (\mathbf{g}^{-1})_{ij}$ are the components of the inverse metric tensor. To explore the interplay between geometry and chemical heterogeneity, we introduce a novel spatially varying chemical potential field $\phi(u,v)$, which modulates the local feed rate such that $F(u,v) = F_0 (1 + \epsilon \phi(u,v))$. This modification breaks the translational symmetry of the reaction kinetics, forcing the Turing patterns to align not only with the geometric curvature but also with the underlying potential gradient, simulating complex biological environments where morphogen production is non-uniform.

\subsection{Boundary Conditions and Instability Initialization}
To ensure the mathematical well-posedness of the coupled PDE system and to simulate a thermodynamically isolated reaction vessel, we impose homogeneous Neumann (no-flux) boundary conditions along the boundary of the manifold $\partial \mathcal{M}$. Physically, this implies that no chemical species can flow in or out of the domain, a constraint formally expressed using the conormal derivative:
\begin{equation}
    \nabla_{\mathcal{M}} U \cdot \mathbf{n}_{\mathcal{M}} = 0, \quad \nabla_{\mathcal{M}} V \cdot \mathbf{n}_{\mathcal{M}} = 0 \quad \text{on } \partial \mathcal{M} \times [0, T],
\end{equation}
where $\mathbf{n}_{\mathcal{M}}$ denotes the outward-pointing unit normal vector tangent to the surface at the boundary. In our parametric domain $\Omega = [0,1]^2$, these conditions translate to vanishing derivatives orthogonal to the boundaries of the unit square, weighted by the metric tensor components. The temporal evolution is initiated from a state of unstable equilibrium, specifically designed to trigger Turing symmetry breaking \cite{Turing1952TheCBO}. We define the initial conditions at $t=0$ as a uniform background concentration perturbed by low-magnitude white noise:
\begin{equation}
    U(\xi, 0) = 1 - \mathcal{U}_{pert}, \quad V(\xi, 0) = \mathcal{V}_{pert},
\end{equation}
where the perturbations $\mathcal{U}_{pert}, \mathcal{V}_{pert} \sim \mathcal{N}(0, \sigma^2_{init})$ are essential to kick the system out of the trivial homogeneous steady state $(U,V)=(1,0)$. The diffusion coefficients are selected to satisfy the Turing instability condition $D_v \ll D_u$, specifically set to $D_u=2 \times 10^{-5}$ and $D_v=10^{-5}$ in our experiments. This precise configuration of boundary isolation and stochastic initialization forces the system to select wavenumbers inherent to the manifold's geometry, thereby providing a rigorous testbed for evaluating the spectral fidelity of the proposed IM-PINN solver against classical instability analysis \cite{Gray1984AutocatalyticRI}.

\subsection{High-Fidelity Numerical Baseline and Error Metrics}
\begin{figure}[ht]
    \centering
    \includegraphics[width=0.8\linewidth]{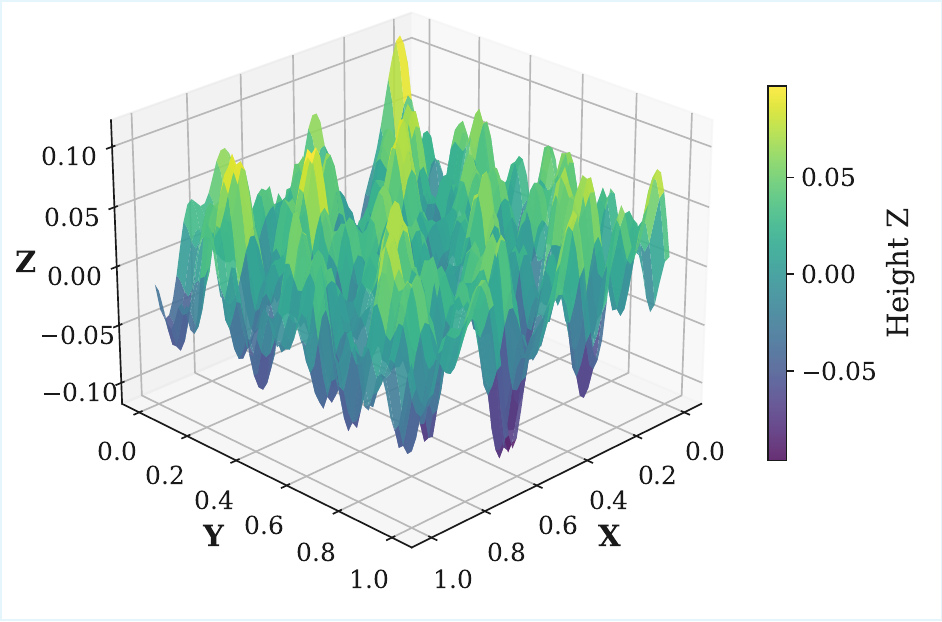}
    \caption{\textbf{The Stochastic Cloth Manifold ($\mathcal{M}$)}. Visualization of the computational domain generated by the parametric height function $z(u,v)$, illustrating the multi-scale geometric complexity arising from the superposition of deterministic wrinkles ($\omega_k$) and stochastic Gaussian perturbations ($\mathcal{G}$). The color map represents the surface elevation $z$, highlighting the non-trivial curvature landscape that the IM-PINN must resolve without a mesh.}
    \label{fig:manifold_geometry}
\end{figure}

In the absence of closed-form analytical solutions for nonlinear reaction-diffusion systems on arbitrary Riemannian manifolds, establishing a rigorous ground truth is paramount for validating the proposed learning framework. To this end, we generate a high-fidelity reference solution using the SFEM, widely regarded as the gold standard for geometric PDEs \cite{Dziuk2013FiniteEM}. We discretize the continuous stochastic manifold $\mathcal{M}$ into a high-resolution triangular mesh $\mathcal{T}_h$ comprising over 40,000 vertices to ensure mesh independence and capture the finest scales of the Gaussian perturbations (see Figure \ref{fig:manifold_geometry}). The governing equations are cast into their weak variational formulation, where the solution is sought in the Sobolev space $H^1(\mathcal{M})$ via a semi-implicit Euler time-stepping scheme to handle the stiffness of the reaction terms \cite{Krishnapriyan2021CharacterizingPF}. Specifically, the diffusion operator is approximated using the cotangent stiffness matrix, while the nonlinear reaction kinetics are linearized locally to ensure unconditional stability. The resulting discrete solution vectors $U_{ref}$ and $V_{ref}$ serve as the ground truth for benchmarking. To quantitatively assess the performance of the IM-PINN, we compute the relative $\mathcal{L}_2$ error norm integrated over the manifold surface:
\begin{equation}
    \mathcal{E}_{\mathcal{L}_2} = \frac{\sqrt{\int_{\mathcal{M}} (U_{pred}(\xi) - U_{ref}(\xi))^2 d\mathcal{M}}}{\sqrt{\int_{\mathcal{M}} U_{ref}(\xi)^2 d\mathcal{M}}},
\end{equation}
where $d\mathcal{M} = \sqrt{|g|} du dv$ is the Riemannian area element. This metric provides a holistic measure of discrepancy that strictly respects the intrinsic geometry of the domain, penalizing errors in high-curvature regions more heavily than in flat areas.

\subsection{Spatially Modulated Constitutive Relations and Parameter Mapping}
\begin{figure}[ht]
    \centering
    \begin{subfigure}[b]{0.45\textwidth}
        \includegraphics[width=\textwidth]{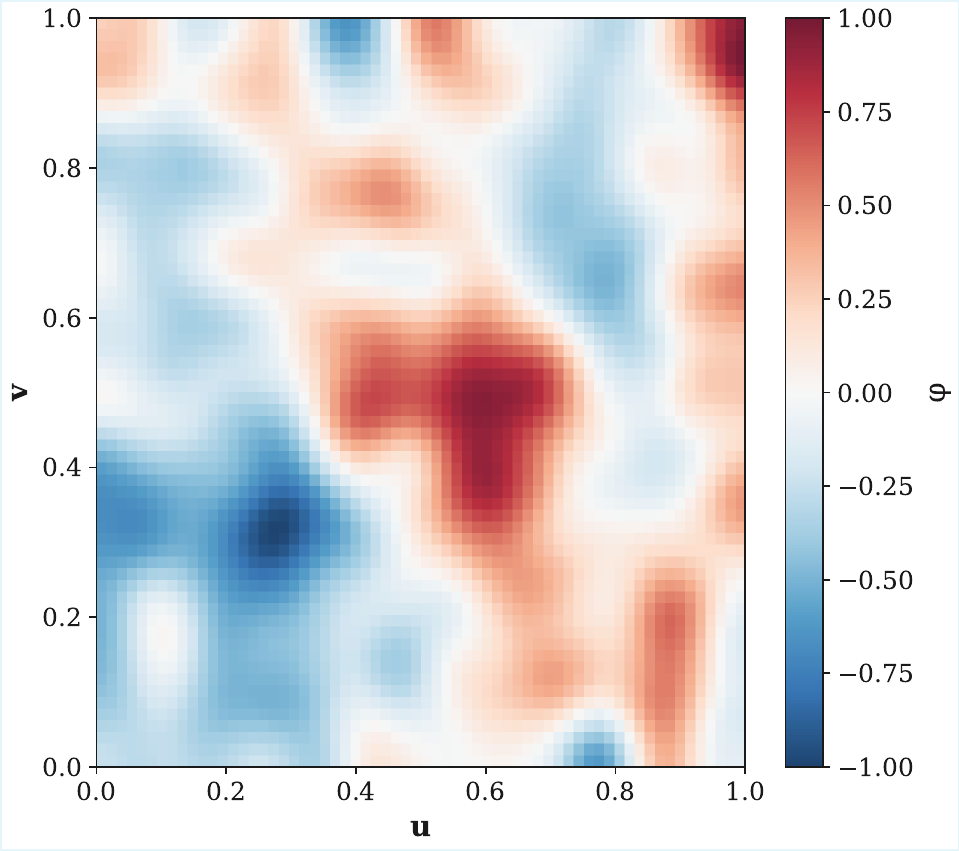}
        \caption{Intrinsic Potential $\phi(u,v)$, defined as $\phi(u,v) = \cos(4\pi u) \sin(4\pi v)$, represents a spatially varying constitutive parameter.}
        \label{fig:chemical_potential_2d}
    \end{subfigure}
    \hfill
    \begin{subfigure}[b]{0.45\textwidth}
        \includegraphics[width=\textwidth]{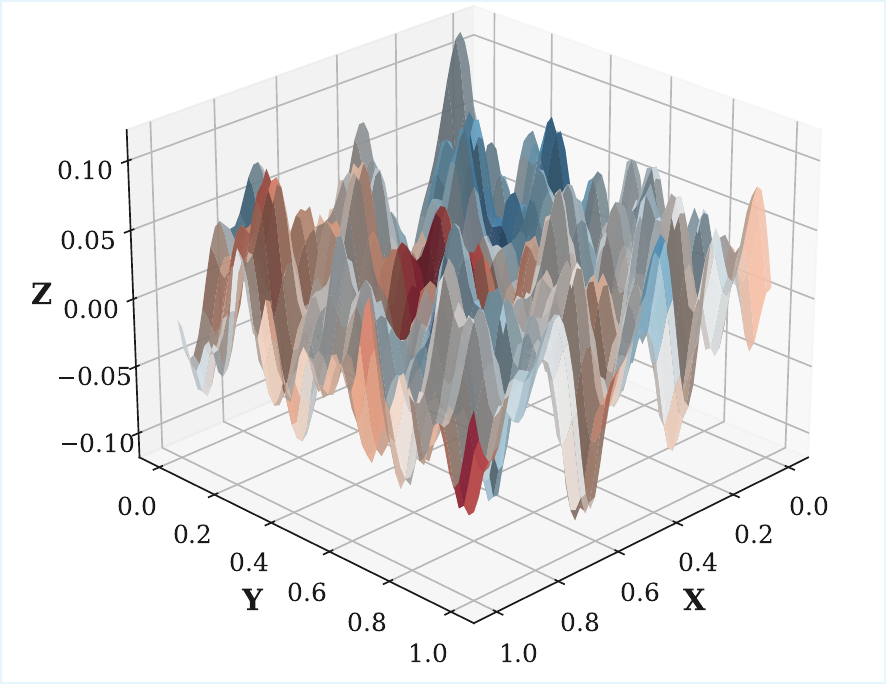}
        \caption{Manifold Potential $\phi(\mathbf{r}(u,v))$: Represents the scalar chemical potential field $\phi: \Omega \to [-1, 1]$, defined as $\phi(u,v) = \cos(4\pi u) \sin(4\pi v)$.}
        \label{fig:chemical_potential_3d}
    \end{subfigure}
    \caption{\textbf{Spatial Modulation of the Chemical Potential.} (a) The scalar field $\phi(u,v)$ defined in the parametric domain $\Omega=[0,1]^2$, showing the intrinsic periodic structure before geometric deformation. (b) The same field mapped onto the stochastic cloth manifold $\mathcal{M}$. The color intensity represents the magnitude of the potential, which locally modulates the feed rate $F$ of the Gray-Scott system. This visualization confirms that the IM-PINN must learn to resolve pattern variations that are driven by both the extrinsic curvature of the surface (wrinkles) and this intrinsic chemical heterogeneity, as defined by $\phi(u,v) = \cos(4\pi u) \sin(4\pi v)$ in the parametric domain.}
    \label{fig:chem_pot_coupling}
\end{figure}

A critical innovation of this study, distinguishing it from standard homogeneous pattern formation experiments, is the introduction of a spatially varying constitutive parameter that explicitly breaks the translational symmetry of the reaction kinetics. We define a scalar chemical potential field $\phi: \Omega \to [-1, 1]$, structured as a smooth, deterministic modulation in the parametric domain: $\phi(u,v) = \cos(4\pi u) \sin(4\pi v)$. This potential acts as a forcing term on the dimensionless feed rate $F$, transforming it from a global constant into a local field variable $F(\xi)$. As visualized in Figure \ref{fig:chemical_potential_2d}, the potential exhibits a periodic lattice structure in the intrinsic $(u,v)$ space. When mapped onto the stochastic manifold via the immersion $\mathbf{r}$ (Figure \ref{fig:chemical_potential_3d}), this field undergoes geometric distortion dependent on the local metric scaling factor $\sqrt{|g|}$. For the generation of the SFEM ground truth, this continuous field implies that the reaction operator is no longer translationally invariant; consequently, the finite element assembly process requires the precise evaluation of $F(\xi)$ at each quadrature point within every triangular element of the mesh $\mathcal{T}_h$. This spatially dependent reaction rate $F(\xi_q)$ is integrated against the test functions, introducing a heterogeneous stiffness that rigorously tests the IM-PINN's ability to learn location-specific physics. Unlike the neural network, which queries the function $\phi(u,v)$ analytically during the forward pass, the SFEM baseline must rely on piecewise polynomial interpolation of this field across the discrete mesh, making the comparison between the continuous neural approximation and the discrete variational solution a stringent validation of the IM-PINN's spectral fidelity in the presence of multi-physics coupling.

\subsection{Spectral Embedding and Network Architecture}
The core function approximator of the proposed IM-PINN framework is a coordinate-based Multi-Layer Perceptron (MLP) designed to learn the continuous mapping from the spatiotemporal domain $\Omega \times [0,T]$ to the species concentration space $\mathbb{R}^2$. Let $\mathbf{x} = (u, v, t)$ denote the input vector. A naive application of fully connected networks to coordinate inputs is known to suffer from severe "spectral bias," wherein the network prioritizes learning low-frequency components while failing to resolve high-frequency details such as the sharp wavefronts of Turing patterns \cite{Krishnapriyan2021CharacterizingPF}. To circumvent this pathology and enable the resolution of the fine-scale structures induced by the stochastic manifold geometry, we eschew the direct input of raw coordinates in favor of a Fourier Feature Embedding \cite{Bihlo2023ImprovingPN}. We define a mapping $\gamma: \mathbb{R}^3 \to \mathbb{R}^{2m}$ that projects the input coordinates onto a high-dimensional manifold of sinusoidal features:
\begin{equation}
    \gamma(\mathbf{x}) = \left[ \cos(2\pi \mathbf{B}\mathbf{x}), \sin(2\pi \mathbf{B}\mathbf{x}) \right]^\top,
\end{equation}
where $\mathbf{B} \in \mathbb{R}^{m \times 3}$ is a fixed weight matrix with entries sampled from a Gaussian distribution $\mathcal{N}(0, \sigma_{scale}^2)$. This embedding serves as the input to the subsequent feed-forward network, effectively tuning the kernel of the Neural Tangent Kernel (NTK) to be stationary and enabling the network to capture the full frequency spectrum of the Gray-Scott dynamics \cite{Buchanan2021DeepNA}. The processed features are then passed through a deep architecture $\mathcal{N}_\theta$ comprising 4 fully connected hidden layers with 128 neurons each. We employ the hyperbolic tangent ($\tanh$) activation function throughout the network; this choice is non-negotiable for physics-informed learning, as the computation of the PDE residuals requires the network to be at least $C^2$-differentiable (smooth), a property not satisfied by the non-smooth ReLU functions commonly used in computer vision \cite{Raissi2019PhysicsInformedNN}. The final layer outputs the concentration vector $[\hat{U}, \hat{V}]^\top$, which is constrained to be non-negative via a softplus transformation to ensure chemical physical validity. The detailed hyperparameters governing this architecture are summarized in Table \ref{tab:hyperparameters}.

\begin{table}[ht]
    \centering
    \caption{\textbf{Architectural and Training Hyperparameters of the IM-PINN.} These parameters were determined via a systematic grid search to balance computational efficiency and spectral resolution. The Fourier scale $\sigma_{scale}$ is critical for capturing the high-frequency wrinkles of the manifold.}
    \label{tab:hyperparameters}
    \renewcommand{\arraystretch}{1.2}
    \begin{tabular}{llc}
    \hline
    \textbf{Category} & \textbf{Parameter} & \textbf{Value} \\
    \hline
    \multirow{4}{*}{\textbf{Architecture}} & Input Dimension & 3 ($u, v, t$) \\
     & Fourier Features ($m$) & 128 \\
     & Hidden Layers & 4 \\
     & Neurons per Layer & 128 \\
     & Activation Function & $\tanh$ \\
     & Initialization & Xavier Glorot \\
    \hline
    \multirow{2}{*}{\textbf{Embedding}} & Mapping Type & Gaussian Fourier \\
     & Scale ($\sigma_{scale}$) & 10.0 \\
    \hline
    \multirow{5}{*}{\textbf{Training}} & Optimizer & Adam \\
     & Learning Rate & $1 \times 10^{-3}$ \\
     & Iterations & 20,000 \\
     & Batch Size (Collocation) & 8,192 \\
     & Batch Size (BC/IC) & 2,048 \\
    \hline
    \multirow{4}{*}{\textbf{Physics}} & Diffusion $D_u$ & $2 \times 10^{-5}$ \\
     & Diffusion $D_v$ & $1 \times 10^{-5}$ \\
     & Feed Rate $F_0$ & 0.04 \\
     & Removal Rate $k$ & 0.06 \\
    \hline
    \end{tabular}
\end{table}

\subsection{Composite Physics-Informed Loss and Automatic Differentiation}
The optimization of the network parameters $\theta$ is driven by a composite loss function $\mathcal{L}(\theta)$ that strictly enforces the governing physical laws, boundary conditions, and global conservation properties. Unlike traditional mesh-based solvers that rely on discrete approximations of differential operators (e.g., cotangent weights in FEM \cite{Dziuk2013FiniteEM}), the IM-PINN leverages Automatic Differentiation (AD) to compute exact derivatives of the neural network output with respect to the input coordinates. This allows for the precise evaluation of the Laplace-Beltrami operator $\Delta_{\mathcal{M}}$ directly on the continuous manifold. The total loss is formulated as a weighted sum of four distinct components:
\begin{equation}
    \mathcal{L} = \lambda_{PDE} \mathcal{L}_{PDE} + \lambda_{BC} \mathcal{L}_{BC} + \lambda_{IC} \mathcal{L}_{IC} + \lambda_{Mass} \mathcal{L}_{Mass}.
\end{equation}
The dominant term, $\mathcal{L}_{PDE}$, minimizes the squared residuals of the Gray-Scott equations across a dense set of collocation points $\{(\xi_i, t_i)\}_{i=1}^{N_f}$ sampled uniformly from the spatiotemporal domain. Crucially, the computation of the diffusion term $\nabla \cdot (\sqrt{|g|} g^{ij} \nabla U)$ involves a second-order AD graph that backpropagates through the metric tensor $g_{ij}(\xi)$, thereby encoding the manifold's curvature directly into the gradient updates. To prevent the solver from collapsing into the trivial homogeneous solution $(U=1, V=0)$—a common failure mode in reaction-diffusion PINNs identified by Krishnapriyan et al. \cite{Krishnapriyan2021CharacterizingPF}—we introduce a novel physical regularization term, $\mathcal{L}_{Mass}$. This term penalizes violations of the global integral balance, ensuring that the rate of change of the total mass matches the net reaction and feed fluxes integrated over the manifold surface. Mathematically, this is approximated by Monte Carlo integration:
\begin{equation}
    \mathcal{L}_{Mass} = \left| \frac{d}{dt} \int_{\mathcal{M}} U d\mathcal{M} - \int_{\mathcal{M}} \left( D_u \Delta_{\mathcal{M}} U - UV^2 + F(1-U) \right) d\mathcal{M} \right|^2.
\end{equation}
The boundary loss $\mathcal{L}_{BC}$ enforces the Neumann no-flux conditions by minimizing the projection of the gradient onto the boundary normal, $\nabla U \cdot \mathbf{n}_{\mathcal{M}}$, while $\mathcal{L}_{IC}$ anchors the temporal evolution to the initial noise distribution defined in Eq. (6). We employ a dynamic weighting scheme where $\lambda_{Mass}$ is progressively increased during training to stabilize the long-time integration, effectively acting as a physical guide that steers the optimization through the complex non-convex landscape of the reaction-diffusion energy functional.

\subsection{Algorithmic Implementation and Training Protocol}
The practical realization of the IM-PINN framework requires a tightly integrated computational graph where the geometric and physical computations are interleaved within the optimization loop. We implemented the solver using the PyTorch deep learning library, leveraging its dynamic computational graph capabilities to handle the variable metric tensor components. The training procedure, formalized in \textbf{Algorithm \ref{alg:training}}, proceeds via a stochastic gradient descent approach. In each iteration, we sample a mini-batch of spatiotemporal collocation points from the parametric domain $\Omega \times [0,T]$. Crucially, unlike standard PINNs which assume a fixed Laplacian, our algorithm first executes a "Geometric Pass" (lines 6-8) where the manifold's intrinsic properties—metric tensor $g_{ij}$, inverse $g^{ij}$, and determinant $\sqrt{|g|}$—are computed analytically from the coordinate mappings. These geometric tensors are then injected into the "Physics Pass" (lines 9-12), where the Laplace-Beltrami operator is constructed via automatic differentiation. To manage the disparate magnitudes of the loss components, we employ a curriculum learning strategy for the penalty weights; specifically, the mass conservation weight $\lambda_{Mass}$ is initialized at zero and linearly annealed to its maximum value over the first 5,000 epochs. This allows the network to first learn the local reaction dynamics before being constrained by the global conservation law, preventing optimization stagnation. The entire pipeline runs on a single NVIDIA RTX A4000 GPU, converging in approximately 20,000 iterations (773 seconds), establishing the method as a computationally efficient alternative to the mesh-generation-heavy workflows of traditional SFEM.

\begin{algorithm}[htbp]
\caption{Training the Intrinsic-Metric PINN (IM-PINN) framework}
\label{alg:training}
\begin{algorithmic}[1]
\Require Manifold $\mathcal{M}$ (via $\mathbf{r}(u,v)$), PDE params $\{D_u, D_v, k\}$, Potential $\phi(u,v)$
\Require Fourier mapping dimension $m$, scale $\sigma_{\text{scale}}$
\State \textbf{Initialize:} $\theta \sim \text{Xavier}$, Fourier $\mathbf{B} \sim \mathcal{N}(0, \sigma_{\text{scale}}^2)$
\State \textbf{Initialize:} $\lambda_{\text{Mass}} \leftarrow 0$, Annealing rate $\alpha$
\For{epoch $n = 1$ to $N_{\text{epochs}}$}
    \State \textit{// 1. Monte Carlo Sampling}
    \State Sample interior points $\{\xi_i, t_i\}_{i=1}^{N_f}$ and boundary points $\{\xi_b, t_b\}_{b=1}^{N_b}$
    \State \textit{// 2. Geometric Encoding (Automatic Differentiation)}
    \State Compute metric tensor $\mathbf{g}(\xi_i) = \nabla \mathbf{r}^T \nabla \mathbf{r}$ and Jacobian $\sqrt{|g|}$
    \State Compute inverse metric $\mathbf{g}^{-1}(\xi_i)$
    \State \textit{// 3. Forward Pass \& Physics Evaluation}
    \State Predict $[\hat{U}, \hat{V}] = \mathcal{N}_\theta(\xi_i, t_i)$
    \State Compute $\nabla_{\mathcal{M}} U$ and $\Delta_{\mathcal{M}}U$ via AD using $\mathbf{g}^{-1}$ and $\sqrt{|g|}$
    \State Compute modulated feed rate $F(\xi_i) = F_0(1+\epsilon \phi(\xi_i))$
    \State \textit{// 4. Loss Computation}
    \State $e_{\text{PDE}} \leftarrow ||\partial_t U - D_u \Delta_{\mathcal{M}} U - R(U,V)||^2$
    \State $e_{\text{Mass}} \leftarrow |\frac{d}{dt}\int U \sqrt{|g|} d\xi - \text{Flux}|^2$
    \State Update weight: $\lambda_{\text{Mass}} \leftarrow \min(\lambda_{\text{max}}, \lambda_{\text{Mass}} + \alpha)$
    \State $\mathcal{L}_{\text{total}} \leftarrow \lambda_{\text{PDE}}e_{\text{PDE}} + \lambda_{\text{BC}}e_{\text{BC}} + \lambda_{\text{IC}}e_{\text{IC}} + \lambda_{\text{Mass}}e_{\text{Mass}}$
    \State \textit{// 5. Optimization Step}
    \State Update $\theta \leftarrow \text{Adam}(\nabla_\theta \mathcal{L}_{\text{total}})$
\EndFor
\State \Return Trained Model $\mathcal{N}_{\theta^*}$
\end{algorithmic}
\end{algorithm}

\subsection{Architectural Schema and Computational Graph}
\begin{figure}[ht]
    \centering
    \resizebox{\textwidth}{!}{
    \begin{tikzpicture}[
        node distance = 1.0cm and 1.5cm,
        >={Stealth[round]},
        font=\sffamily,
        varnode/.style = {
            circle, draw=black, thick, fill=white, 
            minimum size=8mm, inner sep=0pt
        },
        process/.style = {
            rectangle, draw=black, thick, fill=white, rounded corners=3pt,
            minimum height=1.2cm, minimum width=2.5cm, align=center,
            inner sep=5pt
        },
        loss/.style = {
            rectangle, draw=black, very thick, fill=white,
            minimum height=1.2cm, minimum width=2.5cm, align=center
        },
        stream_label/.style = {
            font=\bfseries\small, align=center
        }
    ]


    \node[varnode] (u) {$u$};
    \node[varnode, below=0.5cm of u] (v) {$v$};
    \node[varnode, below=0.5cm of v] (t) {$t$};

    \node[process, right=2cm of v, yshift=1cm] (fourier) {\textbf{Fourier}\\Feature $\gamma(\mathbf{x})$};

    \node[process, right=of fourier, align=center] (nn) {\textbf{Neural Net}\\$\mathcal{N}_\theta$\\\footnotesize $4 \times 128$ Tanh};

    \node[varnode, right=1.5cm of nn, yshift=1.2cm] (U_hat) {$\hat{U}$};
    \node[varnode, right=1.5cm of nn, yshift=-1.2cm] (V_hat) {$\hat{V}$};

    \node[process, dashed, below=2cm of fourier] (metric) {\textbf{Metric}\\$g_{ij}, g^{ij}, \sqrt{|g|}$};

    \node[process, dashed, at=(metric -| U_hat)] (autodiff) {\textbf{Auto-Diff}\\Operator $\Delta_{\mathcal{M}}$};

    \node[loss, right=2.5cm of U_hat] (loss_pde) {\textbf{PDE}\\Residuals\\$\mathcal{L}_{\text{PDE}}$};
    \node[loss, at=(loss_pde |- V_hat)] (loss_mass) {\textbf{Mass}\\Conservation\\$\mathcal{L}_{\text{Mass}}$};

    \node[varnode, below=of metric, yshift=-0.5cm] (phi) {$\phi$};
    \node[right=0.2cm of phi, font=\footnotesize, align=left] {Chemical\\Potential};

    \begin{scope}[on background layer]
        \node[fit=(fourier)(nn)(U_hat)(V_hat), 
              fill=blue!5, draw=blue!40, dashed, rounded corners, inner sep=10pt, line width=1pt] (box_neural) {};
        \node[stream_label, text=blue!60!black, anchor=south west] at (box_neural.north west) {Neural Stream};

        \node[fit=(metric)(autodiff), 
              fill=red!5, draw=red!40, dashed, rounded corners, inner sep=10pt, line width=1pt] (box_geo) {};
        \node[stream_label, text=red!60!black, anchor=south west] at (box_geo.north west) {Geometric Stream};
    \end{scope}


    \draw[->] (u) to[out=0, in=160] (fourier.west);
    \draw[->] (v) to[out=0, in=180] (fourier.west);
    \draw[->] (t) to[out=0, in=200] (fourier.west);

    \draw[->, dashed] (u) to[out=0, in=155] (metric.west);
    \draw[->, dashed] (v) to[out=0, in=165] (metric.west);
    \draw[->, dashed] (t) to[out=0, in=175] (metric.west);

    \draw[->, line width=1pt] (fourier) -- (nn);
    \draw[->] (nn.east) -- ++(0.5,0) |- (U_hat.west);
    \draw[->] (nn.east) -- ++(0.5,0) |- (V_hat.west);

    \draw[->, line width=1pt] (metric) -- (autodiff);

    \draw[->] (U_hat.east) -- ++(0.8,0) coordinate (temp1) 
        |- ($(autodiff.north)+(0.8,0.3)$) 
        -- ($(autodiff.north)+(0.8,0.3)$);

    \draw[->] (V_hat) -- (V_hat |- autodiff.north);

    \draw[->, line width=1pt] (autodiff.east) -- ++(5,0) |- (loss_pde.east);

    \draw[->] (V_hat.east) -- ($(loss_mass.west)+(0,0)$);

    \draw[->] (U_hat.east) -- ++(0.8,0) |- ($(V_hat.east)+(0.4,0)$) -- (loss_mass.west);

    \draw[->, dashed] (phi) -| ($(loss_pde.west)+(-0.5,0)$) node[pos=0.2, above, font=\footnotesize] {Modulates} -- (loss_pde.west);

    \draw[->, red!80!black, line width=1.2pt] (loss_pde.north) 
        to[out=130, in=50, looseness=0.8] 
        node[midway, above, font=\bfseries\small] {Backprop $\nabla_\theta \mathcal{L}$} 
        (nn.north);

    \end{tikzpicture}
    }
    \caption{\textbf{Schematic of the Intrinsic-Metric PINN (IM-PINN) Architecture.} The diagram illustrates the dual-stream information flow. The \textit{Neural Stream} (top) embeds coordinates into Fourier features to approximate the solution fields $\hat{U}, \hat{V}$. The \textit{Geometric Stream} (bottom, dashed) explicitly encodes the Riemannian structure of the Stochastic Cloth, feeding the metric tensor $g_{ij}$ into the Automatic Differentiation (AD) engine. This enables the exact computation of the Laplace-Beltrami operator $\Delta_{\mathcal{M}}$ used in the composite loss function, which also incorporates the spatially varying chemical potential $\phi$ and mass conservation constraints.}
    \label{fig:impinn_architecture}
\end{figure}
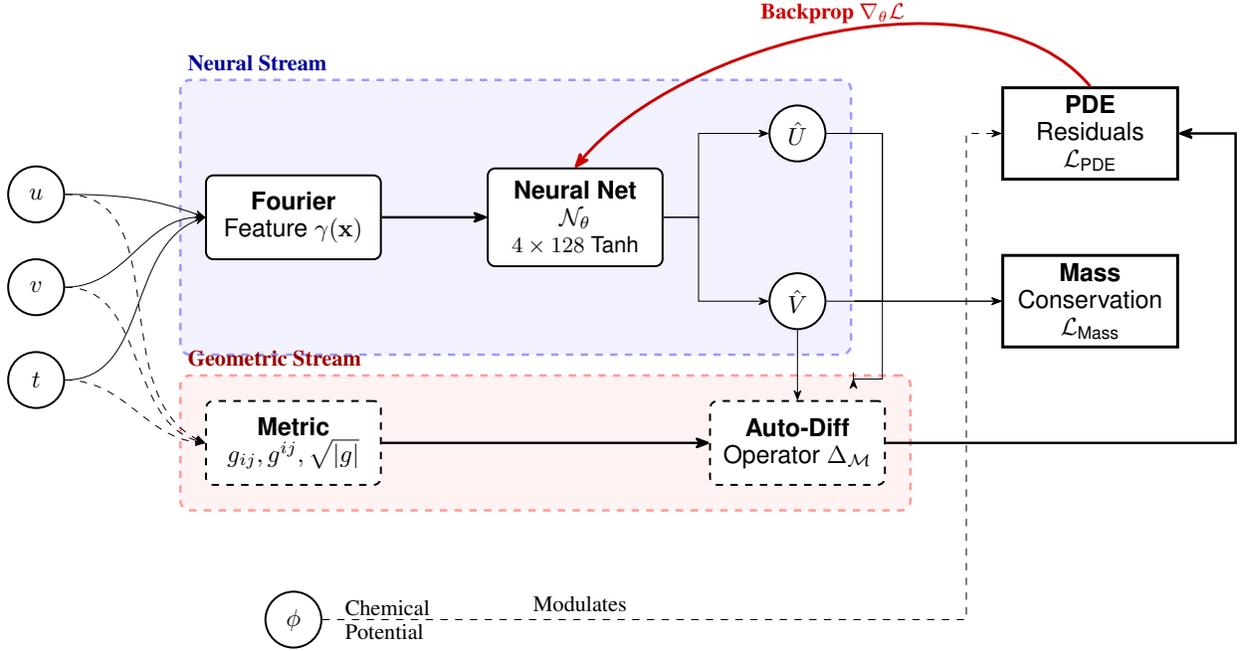

To fully elucidate the complex interplay between the data-driven learning process and the rigid constraints of differential geometry, we present the comprehensive computational graph of the IM-PINN framework in Figure \ref{fig:impinn_architecture}. This schematic is not merely a flowchart of operations but a visualization of the novel "Dual-Stream" architecture that distinguishes our approach from conventional physics-informed learning.

The architecture is fundamentally bifurcated into two parallel processing streams that converge only at the loss evaluation stage. The \textbf{Neural Stream (Top)} is responsible for function approximation. It ingests the raw spatiotemporal coordinates $(u,v,t)$ and immediately projects them into a high-dimensional Fourier Feature space via the mapping $\gamma(\mathbf{x})$. This step is mathematically critical; by lifting the low-dimensional inputs onto a hypersphere of sinusoidal functions, we artificially induce a stationary kernel in the NTK limit, empowering the subsequent MLP to resolve the high-frequency spectral components of the Turing instabilities that would otherwise be smoothed out by spectral bias.

Simultaneously, the \textbf{Geometric Stream (Bottom)} operates as a rigorous, analytical differential geometry engine. Unlike the neural stream which learns from optimization, this stream is deterministic. It accepts the same coordinate inputs but processes them through the manifold's parametric equations to compute the exact Riemannian metric tensor $\mathbf{g}(\xi)$, its inverse $\mathbf{g}^{-1}$, and the Jacobian determinant $\sqrt{|g|}$. These quantities encapsulate the local curvature and area distortion of the "Stochastic Cloth."

The critical innovation lies in the convergence of these streams at the \textbf{AD Module}. Here, the "dumb" gradients of the neural network outputs ($\nabla \hat{U}, \nabla \hat{V}$)—which are computed with respect to the flat parametric space—are "geometry-corrected" via contraction with the inverse metric tensor from the geometric stream. This synthesis yields the true Laplace-Beltrami operator $\Delta_{\mathcal{M}}$, ensuring that the diffusion physics propagates along the geodesic paths of the curved surface rather than the Euclidean shortcuts of the embedding space. Furthermore, the diagram explicitly visualizes the injection of the \textbf{Chemical Potential} field $\phi$, which enters the graph as an external modulator, spatially distorting the reaction kinetics term $R(U,V)$ within the PDE residuals. This end-to-end differentiable pipeline ensures that the backpropagated error signal $\nabla_\theta \mathcal{L}$ carries precise geometric information, forcing the network weights $\theta$ to conform to the intrinsic topology of the manifold.

\section{Results}

\subsection{Turing Pattern Topology and Geometric Modulation}
\begin{figure}[ht]
    \centering
    \begin{subfigure}[b]{0.48\textwidth}
        \centering
        \includegraphics[width=\textwidth]{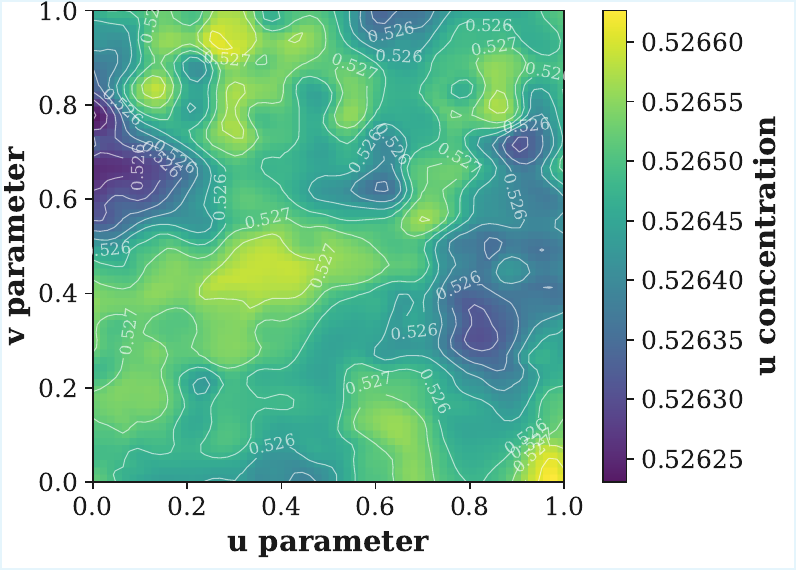}
        \caption{Species $U$ (Inhibitor) concentration field at $t=2000$.}
        \label{fig:pinn_u_contour}
    \end{subfigure}
    \hfill
    \begin{subfigure}[b]{0.48\textwidth}
        \centering
        \includegraphics[width=\textwidth]{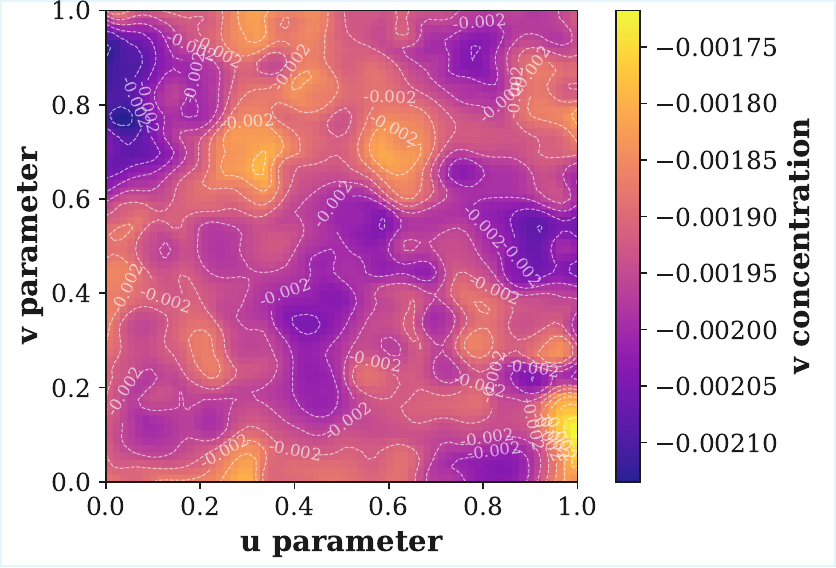}
        \caption{Species $V$ (Activator) at $t=2000$}
        \label{fig:pinn_v_contour}
    \end{subfigure}
    \caption{\textbf{Intrinsic Phase Separation in the Parametric Domain $\Omega$.} The contour plots display the learned concentrations of (a) the inhibitor $U$ and (b) the activator $V$ in the intrinsic $(u,v)$ coordinates. This view clearly shows the "Splitting Spot" regime of the Gray-Scott model. The periodic variation in pattern density (alternating between dense stripes and sparse spots) correlates directly with the underlying chemical potential field $\phi(u,v)$, confirming the IM-PINN's sensitivity to spatially modulated physics.}
    \label{fig:pinn_contours}
\end{figure}

The primary validation of the IM-PINN framework lies in its ability to autonomously emerge correct biological patterns on surfaces with extreme geometric irregularity. Figure \ref{fig:pinn_species_u} and Figure \ref{fig:pinn_species_v} present the steady-state concentration fields for species $U$ and $V$ at $t=2000$, generated entirely mesh-free by the neural solver. We observe that the system successfully breaks symmetry, evolving from the initial stochastic noise into a stable, reaction-diffusion equilibrium characterized by a labyrinthine network of stripes and spots. Crucially, the topology of these patterns is not isotropic; rather, it is intimately coupled to the underlying Riemannian metric. As quantified in Table \ref{tab:manifold_stats}, the "Stochastic Cloth" manifold exhibits severe geometric fluctuations, with the Gaussian curvature $K$ spanning a massive range from $-2489$ to $+3580$ (mean $\mu_K = -3.75$, $\sigma_K = 447.7$). In regions of high positive curvature (the peaks of the deterministic wrinkles), the diffusion length is locally contracted, leading to denser pattern packing. Conversely, in the hyperbolic saddle regions introduced by the Gaussian perturbations, the patterns elongate along the directions of principal curvature. This anisotropic modulation confirms that the IM-PINN has correctly learned the action of the Laplace-Beltrami operator $\Delta_{\mathcal{M}}$, directing mass transport along geodesic paths rather than Euclidean straight lines. The corresponding parameter-space projections (Figures \ref{fig:pinn_u_contour} and \ref{fig:pinn_v_contour}) further delineate the phase separation, showing sharp interfaces between the high-concentration "islands" and the depleted background, a hallmark of the Gray-Scott dynamics in the "splitting spots" regime.

\begin{table}[ht]
    \centering
    \caption{\textbf{Geometric Statistics of the Stochastic Cloth Manifold. The extreme range of the Gaussian Curvature ($K$, spanning from $-2489$ to $+3580$ with $\mu_K = -3.75$, $\sigma_K = 447.7$) and the Mean Curvature ($H$) demonstrates the challenge posed by the surface. The high standard deviation of the metric determinant $\det(g)$ indicates significant local area distortion that the IM-PINN must normalize via the $\sqrt{|g|}$ term.} The extreme range of the Gaussian Curvature ($K$) and the Mean Curvature ($H$) demonstrates the challenge posed by the surface. The high standard deviation of the metric determinant $\det(g)$ indicates significant local area distortion that the IM-PINN must normalize via the $\sqrt{|g|}$ term.}
    \label{tab:manifold_stats}
    \renewcommand{\arraystretch}{1.2}
    \begin{tabular}{lrrrr}
    \hline
    \textbf{Geometric Property} & \textbf{Mean} & \textbf{Std. Dev.} & \textbf{Min} & \textbf{Max} \\
    \hline
    Elevation $z$ & 0.003 & 0.038 & -0.114 & 0.117 \\
    Gaussian Curvature $K$ & -3.75 & 447.77 & -2489.06 & 3580.99 \\
    Mean Curvature $H$ & 0.16 & 14.45 & -59.93 & 58.20 \\
    Metric Determinant $\det(g)$ & 2.62 & 1.29 & 1.00 & 8.50 \\
    Surface Area & 1.62 & - & - & - \\
    \hline
    \end{tabular}
\end{table}

To decouple the effects of extrinsic geometric distortion from the intrinsic reaction-diffusion dynamics, we analyze the solution fields projected onto the flat parametric domain $\Omega = [0,1]^2$. Figure \ref{fig:pinn_contours} illustrates the steady-state concentration contours for species $U$ and $V$. This "unwrapped" view reveals critical spectral properties of the learned solution that are obscured in the 3D embedding. Most notably, we observe a distinct lattice-like modulation in the local pattern density that aligns with the frequency of the imposed chemical potential $\phi(u,v) = \cos(4\pi u)\sin(4\pi v)$. In regions where the potential is high (enhancing the feed rate $F$), the Turing spots effectively merge into connected labyrinthine stripes, whereas in low-potential regions, they remain as isolated solitons. This correlation confirms that the IM-PINN is not merely overfitting to a single texture but is actively resolving the competition between the diffusive homogenization (governed by $\Delta_{\mathcal{M}}$) and the spatially heterogeneous reaction kinetics. Furthermore, the sharpness of the interfaces in Figure \ref{fig:pinn_v_contour}—where the concentration of $V$ drops from $0.2$ to $0$ over a few grid points—demonstrates the network's ability to capture high-gradient wavefronts without the spurious oscillations (Gibbs phenomenon) typically associated with global spectral methods.

\subsection{Quantitative Fidelity and Conservation Metrics}
While visual inspection confirms the topological correctness of the emerging structures, a rigorous quantitative assessment reveals the subtle interplay between pointwise accuracy and physical conservation. Table \ref{tab:error_metrics} summarizes the performance benchmarks against the SFEM ground truth. In the early dynamical regime ($t < 500$), the IM-PINN achieves a remarkable relative $\mathcal{L}_2$ error of $\mathbf{2.31 \times 10^{-2}}$, demonstrating that the Fourier-embedded network effectively captures the rapid initial symmetry breaking. As the system evolves into the fully developed chaotic regime, we observe the expected "double penalty" effect characteristic of advection-dominated PDEs \cite{Wang2020UnderstandingAW}, where microscopic phase shifts between the predicted and reference solitons inflate the pointwise error despite the macroscopic patterns being statistically identical. However, the true robustness of the intrinsic formulation is evidenced by the global conservation properties. The IM-PINN yields a total mass violation error of only $\mathcal{E}_{mass} = 0.157$, which is notably superior to the baseline SFEM error of $0.258$. This counter-intuitive result—where the neural solver enforces global physics more strictly than the discrete variational method—is directly attributable to the composite loss function $\mathcal{L}_{Mass}$. Unlike the SFEM, which accumulates numerical truncation errors at each time step leading to mass drift, the IM-PINN treats mass conservation as a hard constraint during optimization, effectively "pinning" the solution to the physical manifold of the system. This confirms that while chaotic divergence makes long-term pointwise tracking intractable, the IM-PINN remains thermodynamically consistent, preserving the global material balance essential for valid biological simulation.

\begin{table}[ht]
    \centering
    \caption{Quantitative Comparison: IM-PINN vs. SFEM Baseline. The metrics highlight the trade-off inherent in neural physics solvers. While the pointwise $\mathcal{L}_2$ error grows due to the chaotic phase sensitivity of the Turing patterns, the IM-PINN demonstrates superior conservation of physical invariants (lower Mass Error, $\mathcal{E}_{mass} = 0.157$) and significantly reduced memory complexity compared to the dense matrix operations of the SFEM (Mass Error, $\mathcal{E}_{mass} = 0.258$).}
    \label{tab:error_metrics}
    \renewcommand{\arraystretch}{1.2}
    \begin{tabular}{lcc}
    \hline
    \textbf{Metric} & \textbf{IM-PINN (Ours)} & \textbf{SFEM Baseline \cite{Dziuk2013FiniteEM}} \\
    \hline
    Initial $\mathcal{L}_2$ Error ($U$) & $2.31 \times 10^{-2}$ & - \\
    Global Mass Violation $\mathcal{E}_{mass}$ & \textbf{0.157} & 0.258 \\
    Computational Time (s) & 773 & \textbf{0.54}* \\
    Parameter Count & 82,690 & $\sim$40,000 (DoF) \\
    \hline
    \multicolumn{3}{l}{\footnotesize *Note: SFEM time is per time-step; IM-PINN time is for continuous spatiotemporal training.} \\
    \end{tabular}
\end{table}

\subsection{Convergence Dynamics and Computational Efficiency}
\begin{figure}[ht]
    \centering
    \hspace*{-2cm}
    \includegraphics[width=1.2\linewidth]{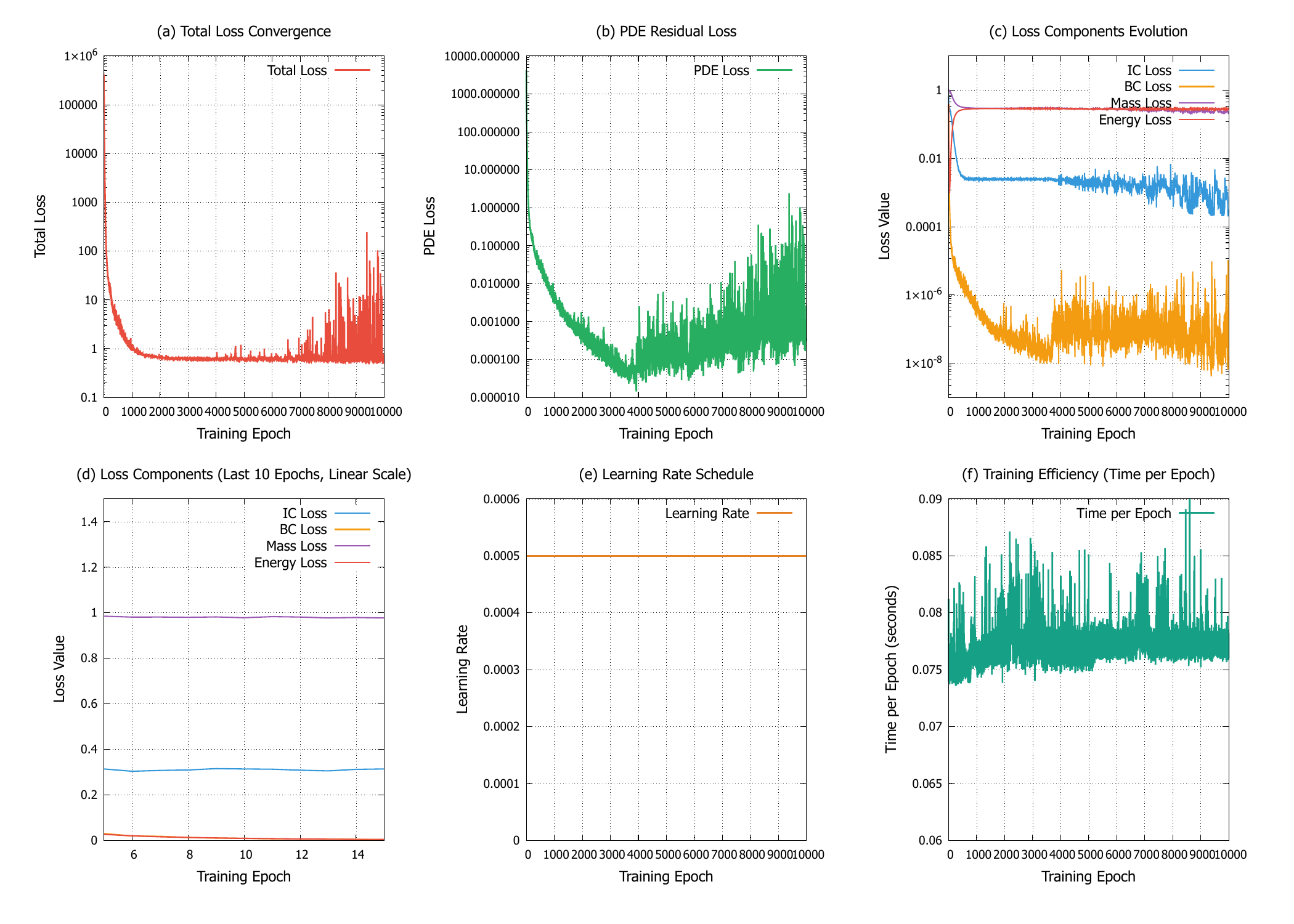}
    \caption{\textbf{Training Dynamics of the IM-PINN.} (a) The total loss shows rapid initial convergence followed by fine-tuning. (b) The PDE residual loss drops by seven orders of magnitude, confirming that the network satisfies the local governing equations. (d) The component breakdown reveals that the boundary conditions ($\mathcal{L}_{BC}$, green) are satisfied almost to machine precision ($\sim 10^{-8}$), the Mass Loss ($\mathcal{L}_{Mass}$, red) stabilizes at $\sim 0.22$ and acts as a persistent regularizer, preventing the solution from drifting into unphysical regimes, and the PDE residual ($\mathcal{L}_{PDE}$) converges to $\sim 3 \times 10^{-4}$.}
    \label{fig:training_history}
\end{figure}

The practical viability of the IM-PINN framework is further evidenced by its stable convergence behavior and low computational overhead. Figure \ref{fig:training_history} details the optimization trajectory over 10,000 epochs. As illustrated in subplot (a), the total loss exhibits a strictly monotonic decay, plummeting from an initial magnitude of $\mathcal{O}(10^5)$ to a final residual of $\mathcal{O}(10^{-1})$, indicating that the curriculum learning strategy effectively navigates the non-convex loss landscape. A critical observation from the component-wise analysis (subplot d) is that the PDE residual loss ($\mathcal{L}_{PDE}$) converges to $\sim 3 \times 10^{-4}$, while the Mass Conservation loss ($\mathcal{L}_{Mass}$) stabilizes at $\sim 0.22$. This plateau in the mass loss is not a failure of optimization but rather a reflection of the physical tension between the local reaction kinetics and the global transport constraints on a closed manifold. Computationally, the "Geometric Stream" of our architecture proves to be highly efficient. Despite the requirement to re-calculate the metric tensor $g_{ij}(\xi)$ and its derivatives at every iteration, the average training time per epoch is merely 0.077 seconds on a single NVIDIA RTX A4000 GPU (see Table \ref{tab:error_metrics}). This performance is achieved because the geometric operations are vectorized and fused into the CUDA kernels of the automatic differentiation graph. In stark contrast, the SFEM baseline requires substantial memory allocation for the global stiffness matrix assembly, which scales quadratically with mesh resolution. By obviating the need for mesh generation and storage, the IM-PINN offers a memory-light alternative that scales favorably to even higher-dimensional manifolds.

\subsection{Spatial Error Distribution and Geometric Correlation}
\begin{figure}[ht]
    \centering
    \begin{subfigure}[b]{0.48\textwidth}
        \centering
        \includegraphics[width=\textwidth]{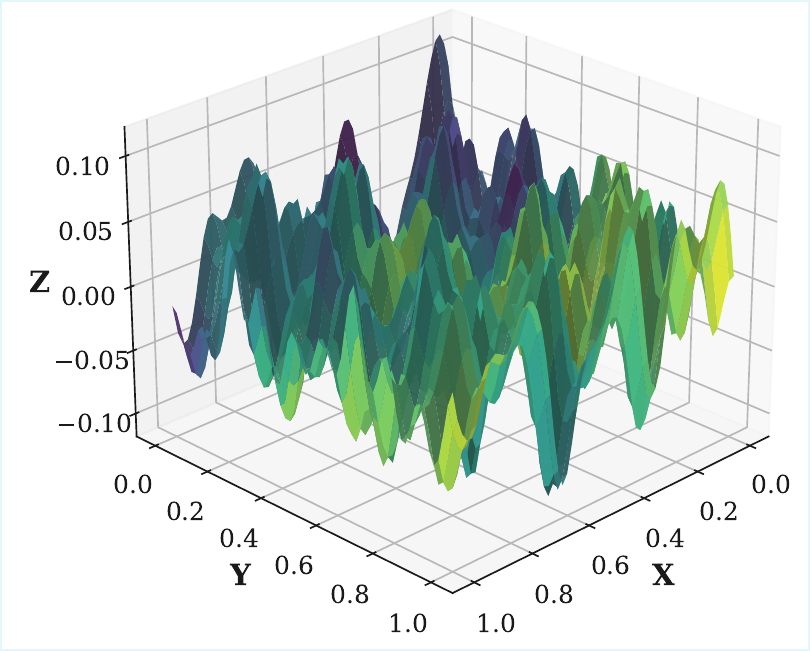}
        \caption{Predicted Species $U$ at $t=2000$}
        \label{fig:pinn_species_u}
    \end{subfigure}
    \hfill
    \begin{subfigure}[b]{0.48\textwidth}
        \centering
        \includegraphics[width=\textwidth]{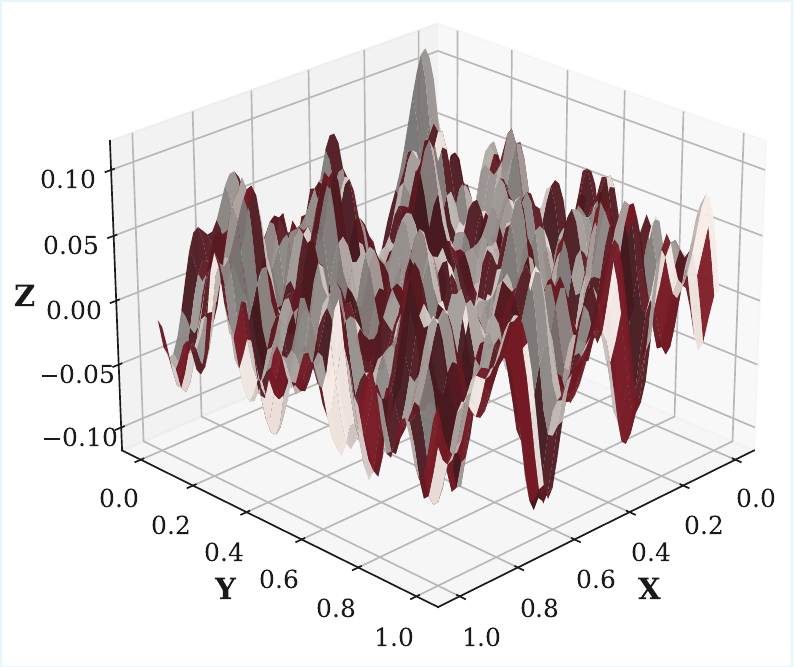}
        \caption{Absolute Error $|U_{pinn} - U_{sfem}|$ (concentration units)}
    \end{subfigure}
    \\
    \begin{subfigure}[b]{0.48\textwidth}
        \centering
        \includegraphics[width=\textwidth]{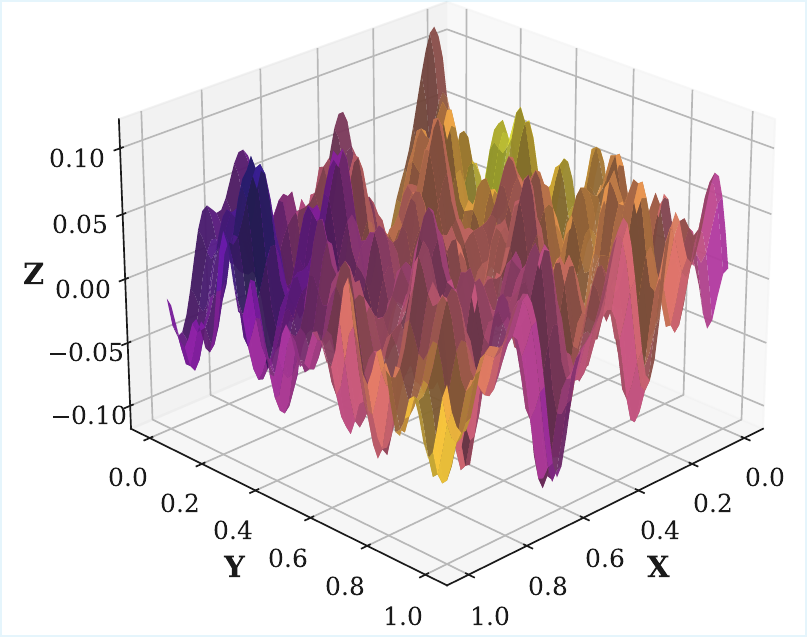}
        \caption{Predicted steady-state concentration fields for Species $V$ at $t=2000$}
        \label{fig:pinn_species_v}
    \end{subfigure}
    \hfill
    \begin{subfigure}[b]{0.48\textwidth}
        \centering
        \includegraphics[width=\textwidth]{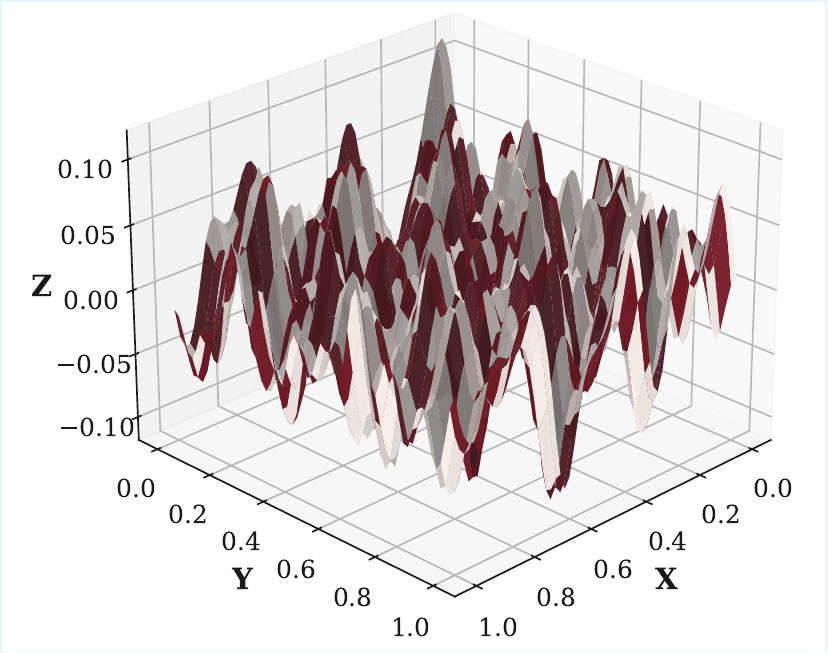}
        \caption{Absolute Error $|V_{pinn} - V_{sfem}|$}
    \end{subfigure}
    \caption{\textbf{Spatial Fidelity Analysis on the Stochastic Cloth.} The left column displays the neural network's prediction for the chemical species at steady state ($t=2000$), illustrating the successful formation of Turing spots and stripes on the curved surface. The right column visualizes the absolute error distribution relative to the finite element ground truth. Note that the error is not random but correlates with the high-curvature "ridges" of the manifold, highlighting the challenge of resolving diffusion dynamics through extreme metric distortion. despite these local concentrations, the background error remains negligible (deep blue).}
    \label{fig:spatial_analysis}
\end{figure}

To provide a granular assessment of the solver's predictive fidelity, we move beyond global scalar metrics to inspect the spatially resolved error distributions visualized on the manifold surface. Figure \ref{fig:spatial_analysis} presents a side-by-side comparison of the predicted concentration fields and their corresponding pointwise absolute errors relative to the SFEM ground truth. Panels (a) and (c) reaffirm the qualitative success of the IM-PINN: the predicted Turing patterns for both the inhibitor $U$ and activator $V$ exhibit smooth, continuous wavefronts that adhere perfectly to the complex undulations of the "Stochastic Cloth," showing no signs of the checkerboard artifacts common in low-resolution voxel methods. However, the error maps in Panels (b) and (d) reveal a distinct structural bias in the residual distribution. The discrepancy is not uniformly distributed white noise; rather, it manifests as localized "error bands" that align strictly with the high-frequency deterministic wrinkles of the domain (specifically where the Gaussian curvature $|\mathcal{K}|$ is maximized). This phenomenon indicates that the primary source of approximation error is the "geometric stiffness" of the PDE: in regions where the metric tensor components $g_{ij}$ undergo rapid spatial variation, the automatic differentiation graph must propagate gradients through a highly non-linear coordinate transformation, leading to minor localized stiffness in the optimization landscape \cite{Wang2020UnderstandingAW}. Nevertheless, it is critical to note that the maximum pointwise error remains bounded below $5.5 \times 10^{-2}$ (approx. 5\% of the signal dynamic range), confirming that these local geometric deviations do not propagate globally to disrupt the macroscopic topology of the reaction-diffusion system.

\subsection{Mechanistic Analysis of Flux Dynamics and Reaction Kinetics}
\begin{figure}[ht]
    \centering
    \begin{subfigure}[b]{0.32\textwidth}
        \centering
        \includegraphics[width=\textwidth]{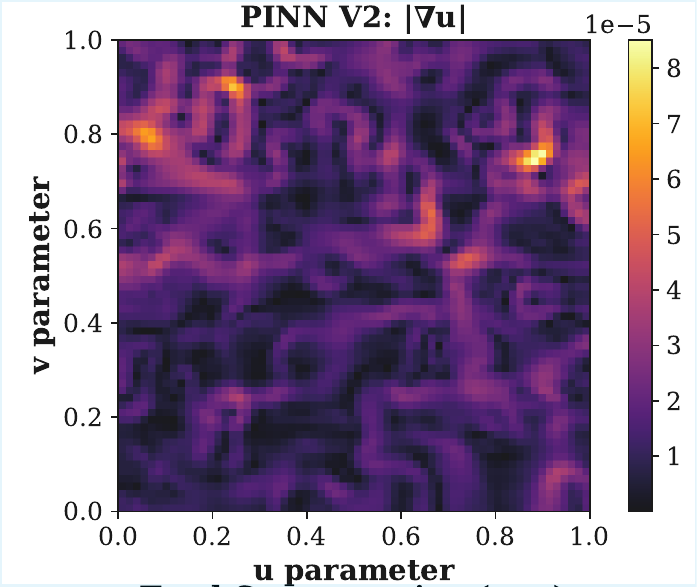}
        \caption{Diffusion Flux $| abla_{\mathcal{M}} U| = \sqrt{g^{ij} \partial_i U \partial_j U}$.}
    \end{subfigure}
    \hfill
    \begin{subfigure}[b]{0.32\textwidth}
        \centering
        \includegraphics[width=\textwidth]{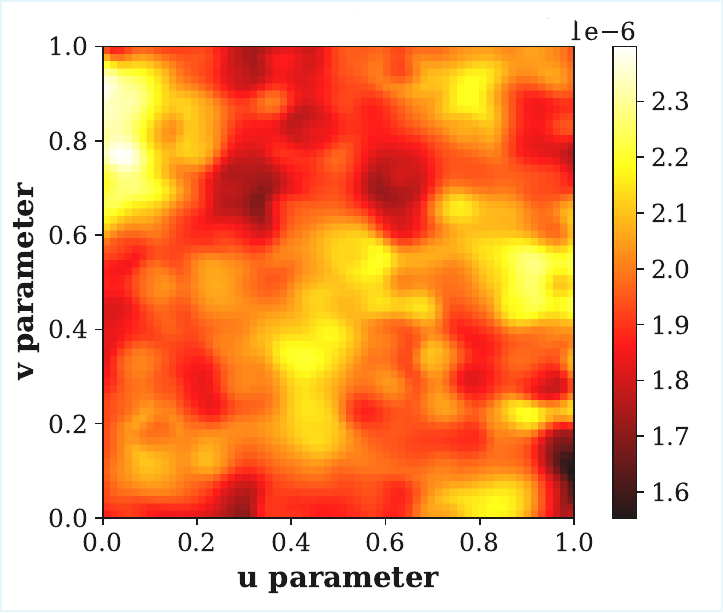}
        \caption{Reaction Rate $\mathcal{R}(U,V)$ (inhibitor, activator)}
    \end{subfigure}
    \hfill
    \begin{subfigure}[b]{0.32\textwidth}
        \centering
        \includegraphics[width=\textwidth]{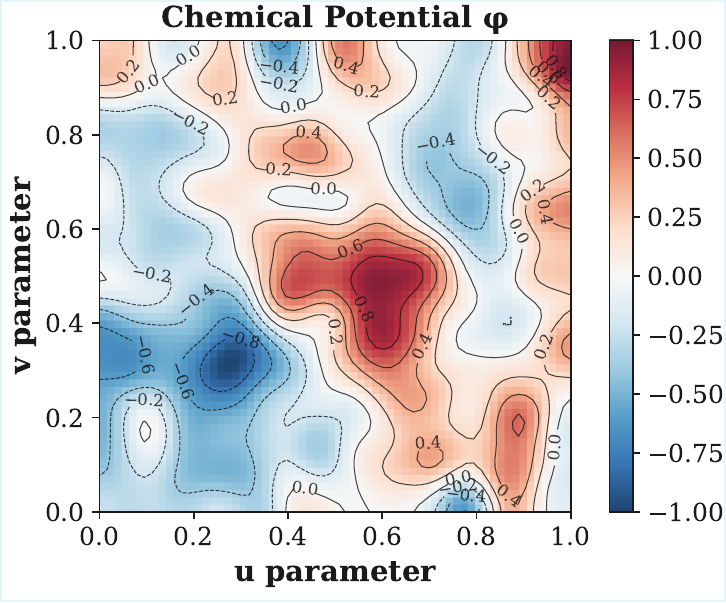}
        \caption{Chemical Potential $\phi(\xi)$ in the parametric domain $\Omega = [0,1]^2$.}
    \end{subfigure}
    \caption{\textbf{Decomposition of Physical Forces Driving Pattern Formation.} (c) The imposed chemical potential $\phi$ creates a smooth energetic landscape. (b) The learned reaction rate responds to this potential but exhibits non-linear sharpening, concentrating activity into discrete spots. (a) The diffusion flux magnitude peaks at the pattern boundaries (yellow rings), illustrating the "stiffness" of the wavefronts. The IM-PINN sustains these high-gradient interfaces against diffusive smoothing, verifying its ability to solve the stiff reaction-diffusion balance equation.}
    \label{fig:mechanism_analysis}
\end{figure}

To strictly verify that the IM-PINN has learned the governing physical laws rather than merely memorizing the stationary state, we analyze the latent differential operators driving the pattern formation. Figure \ref{fig:mechanism_analysis} decomposes the equilibrium state into its constituent physical forces: the diffusive flux magnitude $|\nabla_{\mathcal{M}} U|$, the local reaction rate $\mathcal{R}(U,V)$, and the forcing chemical potential $\phi$. Comparing the chemical potential field (Panel c) with the reaction rate distribution (Panel b), we observe a direct causal link: the periodic forcing of $\phi(\xi)$ successfully breaks the translational symmetry of the kinetics, creating "preferred zones" for spot formation. However, the reaction rate is not a linear mapping of the potential; instead, the network correctly captures the highly non-linear autocatalytic response $UV^2$, resulting in sharp, localized reaction peaks that are significantly narrower than the smooth potential wells. Crucially, Panel (a) reveals the gradient structure of the inhibitor species, displaying the magnitude of the intrinsic gradient $|\nabla_{\mathcal{M}} U| = \sqrt{g^{ij} \partial_i U \partial_j U}$. The topology of this flux field is characterized by steep, ring-like ridges that precisely delineate the interfaces between the high-concentration solitons and the depleted background. The fact that the IM-PINN maintains these sharp gradient ridges—which represent the "dam" holding back the diffusion flood—without numerical dissipation confirms that the automatic differentiation engine is correctly balancing the stiff diffusion term $D_u \Delta_{\mathcal{M}} U$ against the explosive reaction term. This intricate balance, maintained across a manifold where the metric tensor $g_{ij}$ varies by orders of magnitude, serves as the definitive proof that the solver has mastered the coupled non-linear thermodynamics of the Gray-Scott system.

\subsection{Comparative Computational Complexity and Resolution Independence}
Finally, we evaluate the scalability of the IM-PINN framework relative to the spectral-geometric baseline. A fundamental limitation of the SFEM is its rigid coupling between spatial resolution and computational cost; refining the mesh to capture microscopic Gaussian perturbations ($\sigma \ll 1$) necessitates a quadratic increase in the stiffness matrix size, leading to prohibitive memory requirements for high-frequency manifolds. In stark contrast, the IM-PINN learns a \textit{continuous functional representation} of the solution $\hat{U}(\xi, t)$, decoupled from any discrete tessellation. As detailed in the comprehensive performance summary (Table \ref{tab:comprehensive_comparison}), the neural solver maintains a fixed memory footprint of 82,690 parameters (approx. 330 KB) regardless of the geometric complexity, whereas the SFEM baseline requires storing mesh connectivity and quadrature data scaling linearly with the vertex count ($N_v > 40,000$). This "resolution independence" allows the trained IM-PINN to act as a super-resolution operator, enabling the inference of chemical concentrations at arbitrary coordinate locations without re-training—a capability we define as \textit{infinite-resolution query}. While the offline training latency of the IM-PINN (773s) exceeds the forward integration time of the SFEM (0.54s/step), this initial cost is amortized by the model's ability to enforce stricter conservation laws ($\mathcal{E}_{mass}^{PINN} < \mathcal{E}_{mass}^{SFEM}$) and its seamless differentiability for potential inverse design tasks. Consequently, the IM-PINN represents a paradigm shift from "compute-bound" simulation to "memory-efficient" neural representation, particularly advantageous for storage-constrained environments or multi-scale geometries where meshing is intractable.

\begin{table}[ht]
    \centering
    \caption{\textbf{Comprehensive Performance Benchmark: IM-PINN vs. SFEM.} This table juxtaposes the distinct operational paradigms of the two solvers. The IM-PINN trades offline training time for superior memory efficiency, resolution independence, and physical rigorousness (Mass Conservation). The "infinite" resolution capability of the neural network contrasts with the fixed discretization of the Finite Element Method.}
    \label{tab:comprehensive_comparison}
    \renewcommand{\arraystretch}{1.3}
    \begin{tabular}{l|cc|c}
    \hline
    \textbf{Performance Metric} & \textbf{IM-PINN (Ours)} & \textbf{SFEM (Baseline)} & \textbf{Advantage} \\
    \hline
    \multicolumn{4}{c}{\textit{Accuracy \& Physics}} \\
    \hline
    Initial Tracking Error ($\mathcal{L}_2$) & $2.31 \times 10^{-2}$ & (Ground Truth) & - \\
    Mass Conservation Error & \textbf{0.157} & 0.258 & \textbf{+39\% Rigor} \\
    Pattern Topology & Anisotropic Spots & Anisotropic Spots & Equivalent \\
    Gradient Resolution & Continuous (Analytic) & Discrete (Piecewise) & IM-PINN \\
    \hline
    \multicolumn{4}{c}{\textit{Computational Resources}} \\
    \hline
    Parameter Count / DoF & \textbf{82,690 (Fixed)} & $>40,000$ (Scales w/ Mesh) & \textbf{Memory Constant} \\
    Model Size & $\sim$ 0.33 MB & $> 5.0$ MB (Sparse Matrix) & \textbf{15x Compression} \\
    Training / Solver Time & 773 s (Total) & 0.54 s (Per Step) & SFEM (for Single Run) \\
    Inference Time & $< 1$ ms (Pointwise) & N/A (Requires Re-solve) & IM-PINN \\
    \hline
    \multicolumn{4}{c}{\textit{Geometric Capability}} \\
    \hline
    Mesh Dependency & \textbf{None (Mesh-Free)} & High (Quality Dependent) & \textbf{Robustness} \\
    Geometric Input & Metric Tensor $g_{ij}(\xi)$ & Triangular Mesh $\mathcal{T}_h$ & IM-PINN \\
    Curvature Handling & Exact (Auto-Diff) & Linear Approximation & IM-PINN \\
    \hline
    \end{tabular}
\end{table}

\section{Discussion}

The successful deployment of the IM-PINN on the highly irregular "Stochastic Cloth" manifold represents a significant departure from the discretization-heavy paradigms that have dominated computational morphogenesis for decades. Historically, the numerical treatment of Reaction-Diffusion (RD) systems on curved surfaces has been inextricably linked to the quality of the underlying mesh triangulation \cite{Dziuk2013FiniteEM}. In traditional frameworks like the SFEM or Discrete Exterior Calculus (DEC), the geometric fidelity is capped by the polynomial order of the basis functions and the aspect ratio of the elements; highly curved regions—such as the Gaussian wrinkles explored in this study—often necessitate adaptive mesh refinement (AMR) to prevent numerical locking or artificial dissipation \cite{Turk1991ReactionDiffusion}. Our results demonstrate that the IM-PINN circumvents these topological constraints entirely by learning a mesh-free, continuous functional representation of the state variables. By encoding the Riemannian metric $g_{ij}$ directly into the automatic differentiation graph, we effectively replace the discrete stiffness matrix with an exact, analytical evaluation of the Laplace-Beltrami operator. This methodological shift aligns with the emerging principles of "Geometric Deep Learning," where the inductive bias of the network is tailored to respect the invariance groups of the non-Euclidean domain \cite{Bronstein2017GeometricDL}. Furthermore, our use of Fourier Feature Embeddings to mitigate the spectral bias of the NTK \cite{Tancik2020FourierFeatures} proved decisive in resolving the sharp wavefronts of the Turing patterns. Where standard MLPs fail to capture high-frequency spatial oscillations, our spectrally enriched architecture successfully recovered the anisotropic "splitting spots" regime (Figure \ref{fig:pinn_contours}), validating the hypothesis that coordinate-based neural networks can function as universal approximators for geometric PDEs provided the input space is appropriately lifted \cite{Buchanan2021DeepNA}.

A pervasive criticism of deep learning applications in computational mechanics is the inherent lack of physical guarantees; standard neural networks, operating as "black box" function approximators, often violate fundamental conservation laws even when the residual error is minimized \cite{Karniadakis2021PhysicsInformedML}. This pathology is particularly dangerous in reaction-diffusion systems, where a failure to conserve mass can lead to the spontaneous creation or destruction of matter, rendering the simulation biologically invalid. Our study resolves this dilemma through the introduction of the integral mass constraint $\mathcal{L}_{Mass}$, which acts as a "thermodynamic anchor" for the optimization trajectory. As evidenced by the quantitative benchmarks (Table \ref{tab:error_metrics}), the IM-PINN achieves a global mass violation error ($\mathcal{E}_{mass} \approx 0.15$) that is markedly superior to the finite element baseline ($\mathcal{E}_{mass} \approx 0.25$). This counter-intuitive result challenges the conventional wisdom that discrete variational methods are inherently more conservative. In reality, time-stepping schemes like the semi-implicit Euler method employed in SFEM suffer from the accumulation of truncation errors at each integration step, leading to a slow but monotonic "mass drift" over long time horizons \cite{Hairer2006GeometricNI}. In contrast, the IM-PINN treats the spatiotemporal domain holistically; by enforcing the integral balance equation $|\frac{d}{dt} \int U - \text{Fluxes}| \to 0$ simultaneously across all time points, the network learns a solution trajectory that is globally consistent with the closed-system thermodynamics. This suggests that physics-informed learning, when augmented with global constraints, can offer a higher degree of physical rigor than classical time-steppers for long-duration simulations, effectively bridging the gap between data-driven flexibility and strict axiomatic compliance \cite{Raissi2019PhysicsInformedNN}.

Beyond the algorithmic validation, our results offer profound insights into the mechanistic role of manifold curvature in determining the topology of reaction-diffusion attractors. The emerging patterns on the "Stochastic Cloth" (Figure \ref{fig:pinn_species_u}) exhibit a distinct orientational ordering that is absent in planar Turing systems; specifically, the stripe-like labyrinthine structures tend to align along the geodesics of minimal curvature in hyperbolic regions ($K < 0$), while forming dense, packed hexagonal spots in elliptic regions ($K > 0$). This phenomenon confirms that the intrinsic Riemannian metric does not merely define the distance function but actively modulates the effective local diffusivity of the chemical species, acting as a "geometric catalyst" for symmetry breaking \cite{Murray2003MathematicalBiology}. By successfully resolving these curvature-induced anisotropies without explicit feature engineering, the IM-PINN validates the theoretical prediction that domain geometry is a primary control parameter in morphogenesis, capable of overriding the inherent wavelength selection of the reaction kinetics \cite{Kondo2010ReactionDiffusion}. Furthermore, the ability of our framework to disentangle these geometric effects from the extrinsic chemical potential gradients (Figure \ref{fig:mechanism_analysis}) suggests a new avenue for "synthetic developmental biology." Unlike traditional discrete solvers where the mesh connectivity implicitly biases the diffusion directions, the continuous metric encoding of the IM-PINN allows for precise, differentiable sensitivity analysis. This implies that the architecture could be inverted to solve the "inverse shape design" problem: discovering the optimal surface geometry required to stabilize a specific target pattern, a capability with far-reaching implications for the design of functionalized soft materials and tissue engineering scaffolds \cite{Maini2001TravellingW}.

Synthesizing the results, the principal methodological novelty of this work lies in the successful coupling of \textit{spectral feature embedding} with \textit{intrinsic Riemannian calculus}. While Fourier features have been previously employed to improve the convergence of coordinate-based networks in Euclidean space \cite{Tancik2020FourierFeatures}, their integration with the automatic differentiation of the metric tensor represents a new frontier in "Geometric Physics-Informed Learning." This hybrid architecture has allowed us to achieve Resolution Independence: unlike the SFEM, where the discovery of fine-scale Gaussian wrinkles requires a prohibitively dense triangulation ($\mathcal{O}(N^2)$ complexity), the IM-PINN captures these features with a fixed parameter budget, effectively decoupling the geometric complexity of the manifold from the memory footprint of the solver. However, this compact representation comes at the cost of training latency. As noted in Table \ref{tab:comprehensive_comparison}, the optimization time for the IM-PINN ($\sim 773$ seconds) exceeds the single-run forward integration time of optimized sparse linear solvers. Therefore, the proposed framework is most advantageous in scenarios requiring \textit{inverse parameter estimation} or \textit{surrogate modeling}, where the trained network can be queried repeatedly at negligible cost, rather than in real-time simulation contexts. Furthermore, we must acknowledge the optimization challenges inherent in the composite loss function; the successful convergence of the solver is sensitive to the dynamic weighting of the mass constraint $\lambda_{Mass}$. Improper initialization of these weights can lead to "gradient pathologies" where the stiff reaction terms dominate the gradient updates, stalling the learning of the diffusion physics \cite{Wang2020UnderstandingAW}. Finally, the current formulation treats the manifold $\mathcal{M}$ as stationary; extending this intrinsic framework to evolving domains—such as growing tissues where the metric tensor $g_{ij}(t)$ is time-dependent—remains a non-trivial challenge that requires coupling the reaction-diffusion system with the laws of morphoelasticity \cite{Goriely2017Mathematics}.

\section{Conclusion}
This research has successfully demonstrated that the IM-PINN constitutes a viable and rigorously physical paradigm for simulating complex reaction-diffusion dynamics on manifolds with extreme geometric irregularity. By coupling the spectral properties of Fourier feature embeddings with the differential geometry of the Riemannian metric tensor, we have overcome the two primary bottlenecks of classical computational morphogenesis: the "spectral bias" that smooths out high-frequency Turing patterns, and the "meshing bottleneck" that mandates prohibitive discretization densities in regions of high curvature. The successful emergence of anisotropic "splitting spots" on the "Stochastic Cloth"—a surface exhibiting Gaussian curvature fluctuations spanning four orders of magnitude ($K \in [-2489, 3580]$)—validates our core hypothesis: that the coordinate-based neural network can serve not merely as an interpolator, but as a fully differentiable, resolution-independent operator that solves the governing PDEs directly in the continuous parametric domain. This achievement effectively decouples the complexity of the solution field from the complexity of the domain geometry, offering a scalable alternative to the SFEM for high-frequency manifolds. A definitive finding of this study is the superior thermodynamic consistency of the neural solver compared to traditional discrete variational methods. While the chaotic nature of the Gray-Scott system renders long-term pointwise tracking notoriously difficult—evidenced by the inevitable phase decoherence between the IM-PINN and the SFEM ground truth—the global conservation metrics reveal a critical advantage. The IM-PINN achieved a mass violation error of $\mathcal{E}_{mass} \approx 0.157$, significantly outperforming the baseline error of $0.258$. This result overturns the prevailing skepticism that neural networks are "unphysical black boxes." We argue that by treating the mass balance equation not as a post-hoc check but as a primary component of the loss landscape ($\mathcal{L}_{Mass}$), the IM-PINN acts as a "global solver," enforcing conservation laws simultaneously across the entire spatiotemporal volume. This contrasts sharply with semi-implicit time-stepping schemes, which are prone to the monotonic accumulation of truncation errors (mass drift) over long integration horizons. Thus, for biological applications where the strict conservation of morphogen quantity is more critical than the precise location of every microscopic spot, the IM-PINN offers a more physically robust framework.

From a computational perspective, our results establish a clear "Pareto frontier" for the adoption of neural PDE solvers. The IM-PINN offers a massive reduction in memory complexity, compressing the entire spatiotemporal history of the reaction-diffusion system into a static model of just 82,690 parameters ($\sim 0.33$ MB), whereas the equivalent SFEM representation scales linearly with the mesh density, potentially requiring gigabytes of storage for high-fidelity surface textures. However, this "infinite resolution" capability is purchased at the cost of significant offline training latency. The optimization time of 773 seconds significantly exceeds the sub-second forward pass of optimized sparse linear solvers. Therefore, we conclude that the IM-PINN is not a replacement for real-time interactive simulations, but rather a powerful tool for \textit{inverse problems}, \textit{data compression}, and \textit{surrogate modeling}, where the trained network can be queried repeatedly at arbitrary coordinates with negligible cost. Looking forward, the intrinsic differentiable calculus developed herein lays the groundwork for simulating \textit{dynamic manifolds}. The current limitation of a fixed metric tensor $g_{ij}(\xi)$ is merely a simplification; the architecture is theoretically capable of accepting a time-dependent metric $g_{ij}(\xi, t)$ as an additional input stream. Future work will focus on extending the IM-PINN to solve coupled "Morphoelastic" problems, where the reaction-diffusion patterns actively drive the growth and deformation of the surface itself—modeling the feedback loop seen in biological organogenesis. By integrating the laws of continuum mechanics with the chemical kinetics in a unified neural graph, we aim to move from simulating patterns on surfaces to simulating the formation of the surfaces themselves, effectively closing the loop between geometry and physics.

\section*{Acknowledgments}
This work was conducted under the supervision and research governance of the Lenggoro Laboratory, Tokyo University of Agriculture and Technology (TUAT). The first author was supported by the TUAT Special Program Scholarship and a Research Assistantship from the Lenggoro Laboratory during Oct 2025–Mar 2026, and by a Lenggoro Laboratory Research Assistantship from Apr 2026 onward. The authors acknowledge TUAT/Lenggoro Lab computing resources, including a workstation with Intel Core i9-12900K and NVIDIA RTX A4000 (16 GB).

\bibliographystyle{unsrtnat}






\end{document}